\documentclass{article}

\usepackage[preprint,main]{neurips_2025}
\makeatletter
\renewcommand{\@notice}{} 
\makeatother

\usepackage[utf8]{inputenc} 
\usepackage[T1]{fontenc}    
\usepackage{hyperref}       
\usepackage{url}            
\usepackage{booktabs}       
\usepackage{amsfonts}       
\usepackage{nicefrac}       
\usepackage{microtype}      
\usepackage{xcolor}         
\usepackage{caption}
\usepackage{wrapfig}
\usepackage{multirow}
\usepackage{colortbl}
\usepackage{xcolor}
\usepackage{amssymb}
\usepackage{float}
\usepackage{enumitem}
\usepackage[most]{tcolorbox}
\definecolor{purple}{rgb}{.80,.79,.98}
\definecolor{myred}{HTML}{BD463D}
\definecolor{myorange}{HTML}{D38341}
\definecolor{myyellow}{HTML}{DDB405}
\definecolor{mygreen}{HTML}{739B5F}
\definecolor{myblue}{HTML}{6388B5}
\definecolor{mypurple}{HTML}{865FA9}

\usepackage{todonotes}

\title{The Incomplete Bridge: How AI Research (Mis)Engages with Psychology}

\author{%
  Han Jiang\textsuperscript{$\diamondsuit$}\thanks{~Equal contributions.}\ , Pengda Wang\textsuperscript{$\clubsuit$*}, Xiaoyuan Yi\textsuperscript{$\spadesuit$}\thanks{~Corresponding authors.}\ , Xing Xie\textsuperscript{$\spadesuit$},  Ziang Xiao\textsuperscript{$\diamondsuit$$\dagger$ }
  \\
\textsuperscript{$\diamondsuit$}Department of Computer Science, Johns Hopkins University\\
\textsuperscript{$\spadesuit$}Microsoft Research Asia\\
\textsuperscript{$\clubsuit$}Department of Psychological Sciences, Rice University\\
\texttt{hjiang66@jh.edu} \quad \texttt{pw32@rice.edu} \quad \texttt{ziang.xiao@jhu.edu}\\
\texttt{\{xiaoyuanyi,xing.xie\}@microsoft.com} \\
[0.5cm]
}

\begin{document}

\maketitle

\vspace{-15mm}

\begin{figure}[ht]
    \centering
    \includegraphics[width=0.9\linewidth]{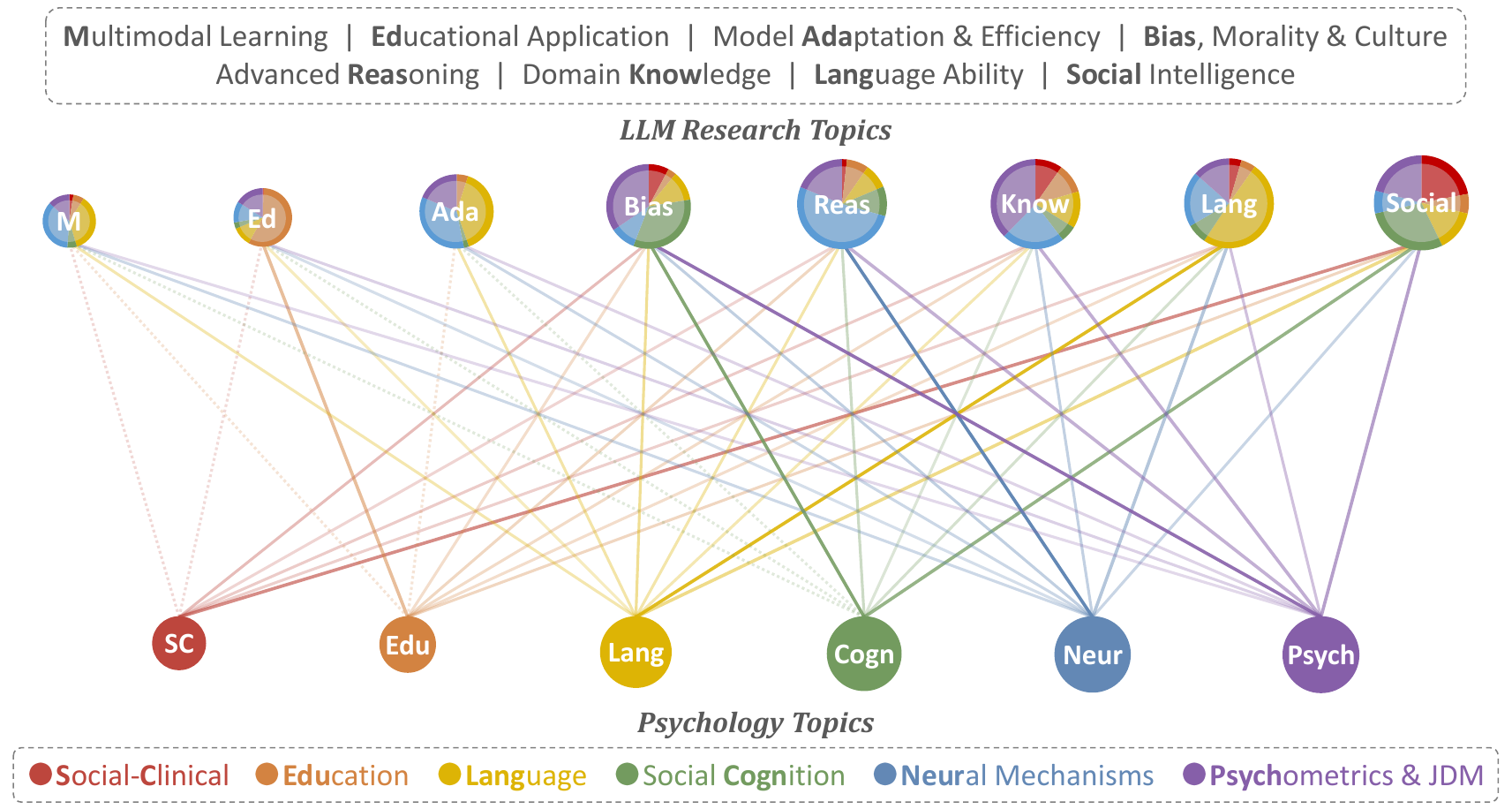}
    \caption{Bipartite Network of Citations from LLM Research Papers to Psychology Papers. \\ \textit{Note:} Pie charts show the distribution of psychology papers (six topics) cited by LLM research papers (eight topics). Circle size indicates the number of papers per topic; line opacity reflects citation frequency; dashed lines represent fewer than ten citations. Abbreviated topic labels are displayed for brevity; complete topic names are provided in §\ref{sec:4}.}
    \label{fig:bipart}
\end{figure}

\begin{abstract}
Social sciences have accumulated a rich body of theories and methodologies for investigating the human mind and behaviors, while offering valuable insights into the design and understanding of Artificial Intelligence (AI) systems.
Focusing on psychology as a prominent case, this study explores the interdisciplinary synergy between AI and the field by analyzing 1,006 LLM-related papers published in premier AI venues between 2023 and 2025, along with the 2,544 psychology publications they cite.
Through our analysis, we identify key patterns of interdisciplinary integration, locate the psychology domains most frequently referenced, and highlight areas that remain underexplored. 
We further examine how psychology theories/frameworks are operationalized and interpreted, identify common types of misapplication, and offer guidance for more effective incorporation.
Our work provides a comprehensive map of interdisciplinary engagement between AI and psychology, thereby facilitating deeper collaboration and advancing AI systems.

\end{abstract}

\clearpage

{\color{black}
\tableofcontents
}

\clearpage

\hypersetup{linkcolor=red}

\section{Introduction}\label{sec:1}

In recent years, the rapid growth of artificial intelligence (AI) has enabled the development of more capable and innovative intelligent systems and has reshaped the way we study and conduct AI research.
One of the most notable trends is toward research pluralism.
Scholars increasingly recognize the importance of complementing traditional AI methodologies with broader approaches to further interpret, guide, and advance contemporary AI systems (e.g.,~\citealp{floridi2023ethics, lin2025fostering}).
As a result, insights from the humanities and social sciences, particularly psychology, linguistics, cognitive science, and philosophy, are being integrated into AI research at an unprecedented scale and depth (e.g.,~\citealp{crawford2021atlas, lake2017building, mccarthy2006proposal}).

One notable example of this interdisciplinary turn is the surge of interest in Large Language Models (LLMs; e.g.,~\citealp{geminiteam2024gemini,meta2025llama4,gpt-4o,gpt-oseries}).
As AI systems progress from functioning primarily as interpreters to also becoming generative agents, LLMs have scaled up from smaller models that originally served as vehicles for language—the principal medium of human communication.
With superior capabilities in natural language understanding and generation, these larger models have demonstrated remarkable performance across a wide range of downstream tasks and are increasingly integrated into practical applications across diverse sectors, including education (e.g.,\citealp{kasneci2023chatgpt}), healthcare (e.g.,\citealp{singhal2023medicine}), law (e.g.,\citealp{katz2024gpt}), scientific research (e.g.,\citealp{meyer2023chatgpt}), and commerce (e.g.,\citealp{li2024ecomgpt}).
In doing so, LLMs have been profoundly redefining the modalities of knowledge acquisition and dissemination (e.g.,~\citealp{gao2024taxonomy}).
However, despite their impressive capabilities, the internal mechanisms of LLMs remain largely opaque (e.g.,\citealp{bommasani2022opportunities}), rendering many of their exhibited behaviors unintended rather than well-understood or explainable (e.g.,\citealp{schaeffer2023emergent, wei2022emergent}).
LLMs are consequently characterized as \textit{`black-box'} systems within both academic and industrial contexts, creating the challenging situation where users expect or steer what occurs without understanding why it occurs (e.g.,~\citealp{lipton2018mythos, rudin2019stop}).
Advancing LLM research urgently requires the development of systematic frameworks for evaluation, interpretability, and human-model interaction, which are essential for addressing foundational challenges related to reasoning mechanisms (e.g.,~\citealp{zhao2024explainability}), capability boundaries (e.g.,~\citealp{10.1145/3641289}), and alignment with human values and safety (e.g.,~\citealp{ouyang2022rlhf}).

This is precisely where interdisciplinary research can play an important role in developing meaningful solutions.
For example, the aforementioned research challenge concerning LLMs parallels a fundamental issue that psychology has grappled with since its inception.
The discipline has progressed through the systematic observation and generalization of intelligent human behavior (e.g.,~\citealp{skinner1965science, watson1913psychology, wundt1904principles}). 
In the absence of direct access to the underlying mechanisms of the human mind, psychologists have historically relied on rigorous experimental designs and theoretical modeling to describe, explain, predict, and influence cognitive and behavioral processes (e.g.,~\citealp{bunge2017philosophy, gigerenzer1991tools, shiffrin1997art}). 
Drawing on this legacy, psychology's empirical tradition and sophisticated experimental paradigms offer a structured, systematic blueprint for advancing LLM research.

However, interdisciplinary collaboration is not without its challenges.
When knowledge is transferred across disciplinary boundaries, researchers often encounter conceptual ambiguities and methodological tensions (e.g.,~\citealp{salter1997outside, huutoniemi2010evaluating}).
Terms may carry different meanings in different fields (e.g., differing interpretations of ``attention'' in LLM and psychology research), which can lead to misunderstandings, misinterpretations, and communication barriers in interdisciplinary research.
Superficial understandings of complex theories may lead to misapplication (e.g., superficial use of motivation theory from psychology to AI design, which ignores human-specific factors like developmental context and lived experience), potentially overlooking critical context-specific factors and resulting in flawed designs or conclusions.
In some cases, speculative or pseudo-scientific reasoning may inadvertently arise (e.g., equating the model’s output diversity with human creativity), which can erode scientific credibility and mislead both academic and public audiences about the nature and capabilities of AI.
Moreover, due to an insufficient or superficial understanding of theory, research may tends to repeatedly draw on a narrow set of well-known theories, favoring familiarity and accessibility over theoretical fit. 
This overreliance can crowd out alternative perspectives, stifle theoretical innovation, and reinforce conceptual blind spots in the field.
All these challenges are deeply felt by AI researchers navigating interdisciplinary work, who often find themselves grappling with questions such as: Which areas of social science are relevant? What theories or frameworks from these areas should we use? How should they be cited appropriately? And how can these concepts be meaningfully and responsibly integrated into technical research?

This paper explores how AI researchers are drawing on psychology literature in their work on LLMs, using this as a way to reflect on how current AI research engages with and integrates interdisciplinary theories and methods.
Specifically, we address the following three core research questions and offer theoretical and methodological recommendations aimed at strengthening research pluralism in AI research, thereby guiding future exploration and facilitating better interdisciplinary research practices:
\begin{itemize}
    \item \textbf{RQ1}: How is psychology research integrated into LLM research?
    \item \textbf{RQ2}: Which psychology theories/frameworks are most commonly used, and which remain underexplored in LLM research?
    \item \textbf{RQ3}: How is psychology research operationalized and interpreted in the context of LLM research?
\end{itemize}

We surveyed 25,843 LLM research articles and compiled a corpus of 1,006 papers that cited psychology research and were published in top-tier AI venues\footnote{NeurIPS, ICLR, ICML, ACL, TACL, EMNLP, and NAACL} between 2023 and 2025. 
From their references, we identified 2,544 psychology papers. 
By analyzing thematic patterns, we mapped the application of psychology research in LLM research, highlighting key areas of interdisciplinary overlap, and revealing potential research gaps.

Building on this foundation, our goal is to provide a rigorous, science-of-science analysis that maps the current intersection of AI and psychology, identifying emerging trends, critical gaps, and opportunities for impactful collaboration. 
By systematically assessing the landscape, we not only help researchers navigate this rapidly evolving space but also highlight areas where psychological insights can meaningfully inform AI development. 
This work is intended to foster responsible, well-informed interdisciplinary research, mitigate risks of conceptual misuse, and ultimately accelerate scientific progress in both fields.
We believe that just as psychology has significantly advanced our understanding of human intelligence, it also holds the potential to play an important role in uncovering and guiding the behavioral mechanisms of AI systems. 
This study represents a key step toward that ambitious vision.

\section{Preliminaries}\label{sec:2}

\subsection{Generative artificial intelligence and large language models}
Generative AI is a subfield of AI that focuses on creating new content, such as text, images, audio, and video~\citep{10.1145/3704262,Feuerriegel2024}, representing a shift from merely interpreting input to generating novel outputs in response to user input.
Among Generative AI systems, LLMs such as GPT~\citep{gpt-4o,gpt-oseries}, Gemini~\citep{geminiteam2024gemini}, and Llama~\citep{meta2025llama4} stand out as prominent text-based technologies. 
Similar progress is also observed in Multimodal Large Language Models~\citep[MLLMs;][]{10.1093/nsr/nwae403}, exemplified by LLaVA~\citep{10655294}, Claude 3~\citep{TheC3}, and GPT-4V~\citep{gpt-4v}, which further support visual content in both input and output.
They have sparked a wave of research spanning their entire life cycle, from architecture and pre-/post-training to application and evaluation, which inherently engages a wide range of human-centered disciplines.

Existing LLMs are primarily decoder-only transformer models~\citep{vaswani2017attention}, in which the attention mechanism (e.g.,~\citealp{NIU202148,Soydaner2022}) is inspired by the concept of selective attention in cognitive science (e.g.,~\citealp{BROADBENT195811,cherry1953some}).
This architecture has greatly benefited from scaling (e.g.,~\citealp{NEURIPS2022_8c22e5e9,kaplan2020scaling}), as first demonstrated by GPT-3~\citep{brown2020language}.
The massive scale places significantly greater demands on computational resources and learning design, which typically occurs in two stages: pre-training, which involves learning from large-scale text corpora and aligns with the goals of corpus linguistics~\citep{hunston2006corpus}; and post-training, which varies depending on specific objectives. Representative post-training methods include: 1) instruction tuning (e.g.,~\citealp{wei2022finetuned,zhang2024instructiontuning}), which enhances LLMs' generalization to unseen tasks;
2) alignment tuning (e.g.,~\citealp{wang2023aligning,wang2024comprehensive}), like Reinforcement Learning from Human Feedback~\citep[RLHF;][]{ouyang2022rlhf} and Direct Preference Optimization~\citep[DPO;][]{rafailov2023dpo}, which aligns LLMs with human intent and preferences;
and 3) Parameter-Efficient Fine-Tuning (PEFT), such as prefix tuning~\citep{li-liang-2021-prefix}, prompt tuning~\citep{lester-etal-2021-power}, and Low-Rank Adaptation~\citep[LoRA;][]{hu2022lora}, which enables effective adaptation of LLMs with a small subset of parameters being updated. 
PEFT is analogous to synaptic plasticity in neuroscience~\citep{Citri2008}, where only specific neural pathways are strengthened or weakened in response to new information.

Since their inception, and beyond language modeling and In-Context Learning~\citep[ICL;][]{dong-etal-2024-survey}, LLMs have demonstrated remarkable advancements in their core capabilities, including long-context modeling (e.g.,~\citealp{liu-etal-2024-lost,yen-etal-2024-long}), advanced reasoning (e.g.,~\citealp{huang-chang-2023-towards,liu2025llmscapablestablereasoning,wei2023cot}), tool use (e.g.,~\citealp{qin2024toolllm}), agency (e.g.,~\citealp{ijcai2024p890,xi2023rise}), and retrieval (e.g.,~\citealp{gao2024retrieval,wang2023augmenting}). 
These emergent capabilities are accelerating their widespread adoption, leading to the development of domain-specific LLMs in fields such as law (e.g.,~\citealp{lai2023largelanguagemodelslaw}), economics (e.g.,~\citealp{horton2023largelanguagemodelssimulated}), medicine (e.g.,~\citealp{singhal2023medicine}), education (e.g.,~\citealp{BERNABEI2023100172}), and the arts (e.g.,~\citealp{wang2024exploringpotentiallargelanguage}). 
Human-AI collaboration is simultaneously strengthening across the broad intersections between LLMs and diverse domains, resulting in many specialized AI assistants that are already in practical use (e.g.,~\citealp{copilot}).
Despite notable progress, concerns about the social risks posed by LLMs continue to be raised (e.g.,~\citealp{bommasani2022opportunities,street2024llm,wang2023decodingtrust,weidinger2021ethical}), making evaluation and alignment crucial steps toward ensuring their responsible use and continuous improvement~\citep{10.1145/3641289,laskar-etal-2024-systematic}. 
Numerous well-crafted benchmarks (e.g.,~\citealp{hendryckstest2021,srivastava2023beyond,10.5555/3666122.3668142,zhong-etal-2024-agieval}) and dynamic approaches (e.g.,~\citealp{jiang2025raising,zhu2024dyval}) have been developed to evaluate LLMs from both holistic and targeted perspectives.
Meanwhile, the objective of alignment is evolving from general human preference (e.g.,~\citealp{ji2023beavertails}) toward personalized preferences (e.g.,~\citealp{salemi-etal-2024-lamp,tan-etal-2024-personalized}), steerability (e.g.,~\citealp{li-etal-2024-steerability,wang-etal-2024-helpsteer}), and social pluralism (e.g.,~\citealp{10.1145/3706598.3713675,feng-etal-2024-modular}), reflecting a deepening exploration of LLMs' social impacts.

Nowadays, these models have long moved beyond being automatons confined to proof-of-concept experiments and are increasingly permeating human society, shaping both everyday life and various industries.
As LLMs approach superhuman performance in certain professional tasks, there is growing interest in their subhuman or possibly non-human-like side. \textit{Do} similar behavioral patterns or principles exist in LLMs~\citep{Gui_2023}? \textit{What} are their internal mechanisms like~\citep{han2025computationmechanismllmposition}? \textit{How} can we learn from human cognition to overcome current limitations in efficiency (e.g.,~\citealp{alizadeh-etal-2024-llm,10589417}), safety (e.g.,~\citealp{10.5555/3692070.3694246,yi-etal-2024-vulnerability}), and social adaptiveness (e.g.,~\citealp{10.1162/opmi_a_00160,LI2025108687})? The intensifying interconnection with human and emergent questions is pushing AI beyond computer science into the broader realm of social sciences.

\subsection{Psychology}
Psychology is an empirical science that systematically studies mental phenomena and behavioral mechanisms. 
It aims to uncover the underlying principles of how individuals perceive, think, feel, and behave in specific contexts (e.g.,~\citealp{colman2016psychology, dewey1892psychology, james1892psychology, schacter2009psychology}). 
Since emerging from philosophy and physiology in the late 19th century, psychology has evolved from an early introspection-based approach to a methodology grounded in empirical research (e.g.,~\citealp{benjamin2023brief, goodwin2015history, schultz2013history}). 
Today, it encompasses a range of subfields, including cognitive, social, developmental, and clinical psychology (see~\citealp{apa_divisions}).

As an interdisciplinary and methodologically diverse science, psychology investigates internal cognitive processes such as attention (e.g.,~\citealp{norman1986attention}), memory (e.g.,~\citealp{atkinson1968human, baddeley2020working}), language (e.g.,~\citealp{chomsky2002syntactic, pinker2003language}), and reasoning (e.g.,~\citealp{doi:10.1126/science.185.4157.1124}), while also exploring how these processes are influenced by emotions (e.g.,~\citealp{damasio2006descartes, ledoux1998emotional}), motivation (e.g.,~\citealp{deci2013intrinsic}, developmental stages (e.g.,~\citealp{piaget1952origins}), and sociocultural environments (e.g.,~\citealp{henrich2010weirdest}).
Researchers employ various methods, such as experimental design (e.g.,~\citealp{reichardt2002experimental}), behavioral observation (e.g.,~\citealp{bakeman2011sequential}), surveys (e.g.,~\citealp{fowler2013survey}), neuroimaging techniques like functional magnetic resonance imaging (fMRI; e.g.,~\citealp{glover2011overview}), and computational modeling (e.g.,~\citealp{guest2021computational}) to study psychological phenomena from multiple dimensions. 
These approaches emphasize the operational definition of variables and statistical inference in order to reveal systematic patterns underlying behavior and mental activity (e.g.,~\citealp{cohen1994earth, kerlinger1966foundations}).

The core objectives of psychology can be delineated into four dimensions (e.g.,~\citealp{coon2013introduction}): description (the systematic observation and documentation of behavior and mental processes), explanation (the elucidation of underlying causes and mechanisms), prediction (the forecasting of future behavior based on theoretical frameworks), and intervention/influence (the ethically grounded facilitation of changes in psychological functioning and behavior).
These objectives exhibit a noteworthy alignment with contemporary inquiries into LLMs. 
While LLMs are constructed through well-defined algorithmic architectures and trained on extensive datasets, many of their sophisticated capabilities (e.g., logical reasoning, code generation) have emerged not as explicit design features, but rather as emergent phenomena associated with increased model scale (e.g.,~\citealp{schaeffer2023emergent, wei2022emergent}). 
Such phenomena underscore the current epistemic gap in our comprehension of LLMs' internal mechanisms: although we are accumulating observations of what these models can do, we still lack a systematic evaluation of their capabilities (e.g.,~\citealp{belinkov2019analysis, bommasani2022opportunities}) and a clear understanding of the underlying reasons for their behaviors (e.g.,~\citealp{10.1145/3641289, zhao2024explainability}).

This epistemological asymmetry naturally invites interdisciplinary engagement. 
In particular, the theoretical paradigms and empirical methodologies developed within psychology may provide a productive lens through which to interrogate and interpret the behavior of LLMs (e.g.,~\citealp{kosinski2023theory, lake2017building}). 
Psychology has historically played an important role in the development of AI, most notably during the early exploration of neural network theory, as exemplified by the perceptron model~\citep{rosenblatt1958perceptron}. 
More recently, psychological insights have continued to inform AI development; for example, attention mechanisms in advanced models (e.g.,~\citealp{vaswani2017attention}) are conceptually inspired by research on human selective attention (e.g.,~\citealp{BROADBENT195811, desimone1995neural, treisman1964selective}).
Furthermore, recent studies have also demonstrated that insights from psychology research can significantly inform and enhance advancements in AI research (e.g., \citealp{liu2025mind,dong2025humanizing,zhang-etal-2024-exploring}).

However, despite this growing body of interdisciplinary work, the broader potential of psychological science to contribute to contemporary AI remains substantial and, in many respects, underexplored.
It is this largely untapped potential that motivates the present study. 
Our goal is to map the intersection between AI and psychology, identify trends, gaps, and opportunities for collaboration, and ultimately advance both fields. 
We hope to help researchers gain a clearer understanding of this domain and promote responsible interdisciplinary collaboration.

\section{Analysis methodology}\label{sec:3}

\subsection{Data collection}\label{sec:3.1}
We began our analysis with papers from seven prominent, peer-reviewed venues in machine learning, artificial intelligence, and natural language processing: 
\begin{itemize}[topsep=-2pt,left=12pt,itemsep=0pt]
    \item \textit{Annual Conference on Neural Information Processing Systems} (NeurIPS)
    \item \textit{International Conference on Learning Representations} (ICLR)
    \item \textit{International Conference on Machine Learning} (ICML)
    \item \textit{Annual Meeting of the Association for Computational Linguistics} (ACL)
    \item \textit{Conference on Empirical Methods in Natural Language Processing} (EMNLP)
    \item \textit{North American Chapter of the Association for Computational Linguistics} (NAACL)
    \item \textit{Transactions of the Association for Computational Linguistics} (TACL)
\end{itemize}
We collected papers presented at the 2023 and 2024 editions of these venues (except for NAACL, a biennial conference, from which we included only the 2024 edition) and added nine papers from Volume 13 of TACL in 2025 (N = 25,843).
To ensure relevance to core LLM research areas, we only included papers that contained the terms \textit{LLM} or \textit{language model} in their title or abstract (N = 3,962). 

Subsequently, we extracted citation data for the remaining papers using the Semantic Scholar Academic Graph (S2AG) API\footnote{https://www.semanticscholar.org/product/api}, which indexes over 214 million scholarly works across diverse scientific domains. 
From this citation dataset, we identified and retained only those referenced papers classified under the field of \textit{Psychology} but not under \textit{Computer Science}, according to Semantic Scholar’s disciplinary tagging.
LLM-related papers that did not cite at least one such psychology reference were excluded from the final sample (N = 1,006).

Following this multi-step curation process, our final dataset comprised 1,006 LLM research papers and 2,544 cited psychology reference papers.

\subsection{Embedding and clustering}\label{sec:3.2}
We employed the K-means clustering algorithm~\citep{k-means,1056489,macqueen1967some} to discern thematic groupings within corpora of LLM research papers and psychology reference papers, respectively.
Specifically, we used the SPECTER model~\citep{cohan-etal-2020-specter} to generate embeddings for each paper. 
SPECTER is a transformer model trained on citation networks to produce document-level embeddings; it takes the title and abstract of a paper as input.
Clustering was then performed across a range of cluster counts \( K \in [4, 10] \), with the silhouette coefficient computed for each configuration to assess clustering quality. 
This procedure was repeated 50 times, and the value of \( K \) that yielded the highest average silhouette coefficient was selected as optimal.

This process yielded eight clusters for the LLM research papers and six for the psychology papers. 
The topic of each cluster was then inferred through a two-stage process: first, by summarizing the paper titles and abstracts within each cluster into five salient phrases using GPT-4o across ten runs, and second, by manually synthesizing these outputs into a concise, representative cluster label. The instruction template for summarization is provided in App.~\ref{app:instruction}, and the complete cluster names and descriptions can be found in §\ref{sec:4}.

\subsection{Psychology theory/framework extraction and connection}\label{sec:3.3}
To identify popular psychology theories and frameworks studied in the 2,544 psychology reference papers, we conducted the following three-step process:
First, following the practice described in §\ref{sec:3.2}, we clustered the psychology papers within each of the six aforementioned clusters, resulting in 32 secondary clusters.
Next, in each secondary cluster, we 1) randomly sampled 20 papers and summarized their titles and abstracts into five key phrases using GPT-4o, which was repeated ten times; 
and 2) hired domain experts to derive a cluster label, i.e., a research topic, and identify two to four primary psychology theories or frameworks based on the summaries from GPT-4o.
Additionally, the experts were invited to suggest three more psychology theories or frameworks that are well-known in psychology but under-explored in LLM research for each primary cluster.
Finally, we got a total of 96 popular psychology theories and frameworks, which were then linked to both psychology and LLM research papers.

To connect the identified theories and frameworks with the psychology papers in each primary cluster, we provided GPT-4.1 with the title and abstract of a paper, along with the list of popular theories/frameworks from a secondary cluster in each query. GPT-4.1 was then asked to determine whether the paper involves any of the listed theories or frameworks. The instruction template for the relevance judgment is provided in App.~\ref{app:instruction}.
Once the relevant psychology papers were identified, we considered an LLM research paper to be associated with a given theory or framework if it cited any of the corresponding psychology papers.
In this way, we calculated the citation count for each identified psychology theory or framework across the surveyed papers. For each primary cluster, the three most frequently cited psychology theories/frameworks and the three underexplored ones were selected for analysis in §\ref{sec:5.2}. The full list of secondary cluster names and the extracted popular psychology theories/frameworks are shown in Tables \ref{fig:theory_0}-\ref{fig:theory_5}.

\section{Clustering structure}\label{sec:4}

\begin{figure}[ht]
    \centering
    \includegraphics[width=\linewidth]{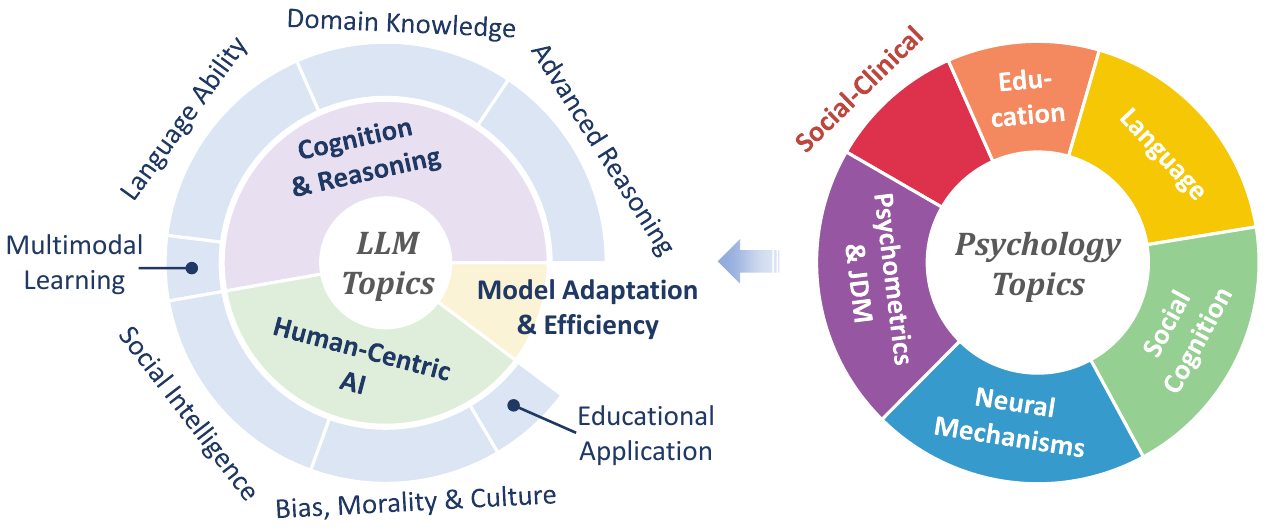}
    \caption{Illustration of Eight LLM Research Clusters and Six Psychology Clusters.\\ \textit{Note:} The central angle of each segment indicates the proportion of research papers associated with its corresponding cluster. The LLM research clusters are organized in a two-tiered structure, with \textit{Cognition \& Reasoning}, \textit{Human-Centric AI}, and \textit{Model Adaptation \& Efficiency} constituting the primary thematic layer. Abbreviated cluster labels are displayed for brevity; complete cluster names are provided in §\ref{sec:4}.}
    \label{fig:clusters}
\end{figure}

As we mentioned in the previous section, we derive eight distinct LLM research clusters (shown in Fig.~\ref{fig:clusters}, left) and six distinct psychology clusters (shown in Fig.~\ref{fig:clusters}, right). 
In this section, we present the names of the identified clusters and offer a brief overview of each.
We begin with the LLM research clusters, followed by the psychology research clusters. 
For both sets, the corresponding topics are listed in ascending order based on the number of papers associated with each cluster.

\subsection{LLM research clusters}

\textbf{1)\enspace Multimodal Comprehension and Spatial Reasoning}
\begin{itemize}[topsep=-2pt,left=12pt,itemsep=0pt]
    \item \textit{Abbreviated as \textbf{Multimodal Learning}}
    \item This cluster is characterized by the integration of modalities beyond text, such as images~(e.g., \citealp{chakrabarty-etal-2023-spy}), audio~(e.g., \citealp{10.5555/3692070.3693076}), video~(e.g., \citealp{10.5555/3737916.3741129}), and time series~(e.g., \citealp{jin2023time}). 
    Building on early research that primarily leveraged LLMs for the textual component of existing multimodal tasks, later directions including spatial reasoning~(e.g., \citealp{wu2024minds}), concept binding~(e.g., \citealp{li-etal-2024-cognitive}), and multimodal generation~(e.g., \citealp{10.5555/3692070.3694603}) have emerged with advances in LLMs and MLLMs. 
    This branch of research has laid the foundation for embodied AI and more real-world applications. 
\end{itemize}

\vspace{0.5em}
\textbf{2)\enspace Educational Applications and Pedagogical Alignment}
\begin{itemize}[topsep=-2pt,left=12pt,itemsep=0pt]
    \item \textit{Abbreviated as \textbf{Educational Application}} 
    \item This cluster explores how LLMs can be applied in educational settings, including educational material generation~(e.g., \citealp{luo-etal-2024-chain}), assessment methods~(e.g., \citealp{xiao-etal-2023-evaluating-evaluation}), instructional design~(e.g., \citealp{yin-etal-2023-read}), and intelligent tutoring systems~(e.g., \citealp{sonkar-etal-2023-class}).
    The goal is to align models with sound pedagogical principles and ensure their effectiveness in supporting human teaching and learning. 
\end{itemize}

\vspace{0.5em}
\textbf{3) Scalable and Efficient Algorithms for Learning and Inference}
\begin{itemize}[topsep=-2pt,left=12pt,itemsep=0pt]
    \item \textit{Abbreviated as \textbf{Model Adaptation \& Efficiency}}
    \item This cluster aims to improve the scalability and efficiency of LLM adaptation methods, encompassing pre-training~(e.g., \citealp{10.5555/3692070.3692457}), post-training~(e.g., \citealp{munos2024nash}), and inference-time adaptation~(e.g., \citealp{zhang2023ho}).
    The emphasis is on trade-offs between various aspects of the learning algorithms, such as overall performance versus computational cost~(e.g., \citealp{10.5555/3666122.3666563}) and alignment performance versus pre-training capabilities~(e.g., \citealp{lin-etal-2024-mitigating}). 
    In general, it focuses on relatively low-level algorithm designs and serves to accommodate a variety of expectations and use cases.
\end{itemize}

\vspace{0.5em}
\textbf{4) Bias Measurement, and Moral and Cultural Alignment and Evaluation}
\begin{itemize}[topsep=-2pt,left=12pt,itemsep=0pt]
    \item \textit{Abbreviated as \textbf{Bias, Morality \& Culture}}
    \item This cluster mainly addresses bias in LLMs~(e.g., \citealp{10.5555/3692070.3693479}), which occurs as a consequence of complex interactions among morality~(e.g., \citealp{abdulhai-etal-2024-moral,scherrer2023evaluating}), culture~(e.g., \citealp{li2024culturellm,shen-etal-2024-understanding}), ideology~(e.g., \citealp{plaza-del-arco-etal-2024-divine}), and other factors. 
    This line of research seeks to measure and mitigate harmful stereotypes by decomposing them into different social aspects and conducting analysis and alignment within each, so that LLMs can better respect diverse moral frameworks and cultural lenses during interaction.
\end{itemize}

\vspace{0.5em}
\textbf{5) Advanced Reasoning and Theory of Mind in Multi-Agent Systems}
\begin{itemize}[topsep=-2pt,left=12pt,itemsep=0pt]
    \item \textit{Abbreviated as \textbf{Advanced Reasoning}}
    \item This cluster explores high-level reasoning abilities~(e.g., \citealp{huang-chang-2023-towards}) that emerge with the upscaling of LLMs, including logical reasoning~(e.g., \citealp{wang2024hypothesis}), mathematical reasoning~(e.g., \citealp{imani-etal-2023-mathprompter}), and planning~(e.g., \citealp{valmeekam2023on}). Another prominent subfield is theory of mind in multi-agent scenarios~(e.g., \citealp{li-etal-2023-theory,wu-etal-2023-hi}), which enables LLMs to infer others’ mental states—an ability essential for collaborative and socially intelligent systems. However, whether LLM reasoning constitutes merely structured, goal-directed pattern completion or resembles human-like thinking remains an open question.
\end{itemize}

\vspace{0.5em}
\textbf{6) Knowledge Utilization and Domain-Specific Applications}
\begin{itemize}[topsep=-2pt,left=12pt,itemsep=0pt]
    \item \textit{Abbreviated as \textbf{Domain Knowledge}}
    \item This cluster enhances the ability of LLMs to manage and utilize knowledge, including resolving knowledge conflicts~(e.g., \citealp{xu-etal-2024-knowledge-conflicts}), performing knowledge-grounded reasoning~(e.g., \citealp{chen-etal-2024-temporal}), and conducting fact verification~(e.g., \citealp{pan-etal-2023-risk}). 
    Once the faithfulness of the information is ensured, the processed knowledge, both structured and unstructured, can be applied across domains such as medicine~(e.g., \citealp{kim2024mdagents}), law~(e.g., \citealp{fei-etal-2024-lawbench}), and other areas where factual accuracy and specialized understanding are critical for practical applications.
\end{itemize}

\vspace{0.5em}
\textbf{7) Linguistic Competence, Multilingual Adaptation, and Text Generation Quality}
\begin{itemize}[topsep=-2pt,left=12pt,itemsep=0pt]
    \item \textit{Abbreviated as \textbf{Language Ability}} 
    \item This cluster focuses on the core capability of LLMs—language ability. Research primarily investigates basic linguistic processing~(e.g., \citealp{kobayashi-etal-2024-revisiting}) and multilingual understanding~(e.g., \citealp{tang-etal-2024-language,zhang-etal-2023-multilingual,zhang-etal-2023-dont}), as well as more advanced language phenomena such as analogy~(e.g., \citealp{wijesiriwardene-etal-2023-analogical}), creativity~(e.g., \citealp{gomez-rodriguez-williams-2023-confederacy}), metaphor~(e.g., \citealp{joseph-etal-2023-newsmet,wachowiak-gromann-2023-gpt}), and ellipsis~(e.g., \citealp{hardt-2023-ellipsis,testa-etal-2023-understand}). 
    It aims to produce outputs that are grammatically correct, semantically coherent, and contextually appropriate.
\end{itemize}

\vspace{0.5em}
\textbf{8) Socially Aware and Emotionally Intelligent Dialogue Systems}
\begin{itemize}[topsep=-2pt,left=12pt,itemsep=0pt]
    \item \textit{Abbreviated as \textbf{Social Intelligence}}
    \item This cluster centers on the social adaptiveness of LLMs—the ability to understand and navigate social situations effectively. 
    An intelligent system should first avoid producing harmful content~(e.g., \citealp{shaikh-etal-2023-second,wei2023jailbroken}), then develop an understanding of diverse social dynamics~(e.g., \citealp{zhao-etal-2024-large,zhou2024sotopia}), enabling it to engage appropriately in social interactions~(e.g., \citealp{kwon-etal-2024-llms,shao-etal-2023-character}) and deliver emotionally resonant responses~(e.g., \citealp{chen-etal-2023-soulchat,sabour-etal-2024-emobench}), thereby fostering beneficial relationships between humans and AI in society.
\end{itemize}

\subsection{Psychology research clusters}

\textbf{1) Social-Clinical Psychology of Mental Health and Intervention}
\begin{itemize}[topsep=-2pt,left=12pt,itemsep=0pt]
    \item \textit{Abbreviated as \textbf{\textcolor{myred}{Social-Clinical}}}
    \item This cluster explores the psychological foundations of mental health and clinical practice. 
    It includes research on social influences (e.g., \citealp{liao2020misunderstood,meyer2003prejudice}), therapeutic interventions (e.g., \citealp{fitzpatrick2017delivering,greimel2011cognitive}), and the psychological processes that underlie well-being (e.g., \citealp{diener1985satisfaction,diener2010new}), stress (e.g., \citealp{lazarus1966psychological,spitzer2006brief}), and disorder (e.g., \citealp{cuijpers2010self,persson2019revisiting}). 
\end{itemize}

\vspace{0.5em}
\textbf{2) Learning, Teaching Design, and Educational Development}
\begin{itemize}[topsep=-2pt,left=12pt,itemsep=0pt]
    \item \textit{Abbreviated as \textbf{\textcolor{myorange}{Education}}}
    \item This cluster focuses on how people learn and how educational environments can be optimized. 
    It investigates instructional strategies (e.g., \citealp{kirschner2006minimal,miri2007purposely}), developmental pathways (e.g., \citealp{stipek1989developmental,zimmerman2000development}), and the cognitive mechanisms that support effective learning (e.g., \citealp{garner1987metacognition,pintrich2002role}) and teaching (e.g., \citealp{kraft2018effect,sullivan2014use}). 
\end{itemize}

\vspace{0.5em}
\textbf{3) Language Comprehension, Pragmatic, and Psycholinguistic}
\begin{itemize}[topsep=-2pt,left=12pt,itemsep=0pt]
    \item \textit{Abbreviated as \textbf{\textcolor{myyellow}{Language}}}
    \item This cluster examines the psychological and cognitive processes involved in understanding and using language. 
    Topics include real-time language comprehension (e.g., \citealp{ehrlich1981contextual,levy2008expectation}), pragmatic inference (e.g., \citealp{goodman2016pragmatic,levinson2000presumptive}), and the development (e.g., \citealp{berko1958child,oates2004cognitive}) and disorders (e.g., \citealp{boschi2017connected,gorno2011classification}) of language. 
\end{itemize}

\vspace{0.5em}
\textbf{4) Emotion, Morality, and Culture in Social Cognition}
\begin{itemize}[topsep=-2pt,left=12pt,itemsep=0pt]
    \item \textit{Abbreviated as \textbf{\textcolor{mygreen}{Social Cognition}}}
    \item This cluster investigates how emotions, moral reasoning, and cultural context shape our social understanding. 
    It includes research on emotions (e.g., \citealp{moors2013appraisal,scherer2019emotion}), empathy (e.g., \citealp{hoffman1996empathy,konrath2018development}), value systems (e.g., \citealp{schwartz2012overview,graham2013moral}), identity (e.g., \citealp{hegarty2018nonbinary,roccas2002social}), and the ways people perceive and interact with others (e.g., \citealp{brown1986evaluations,cuddy2009stereotype}). 
\end{itemize}

\vspace{0.5em}
\textbf{5) Neural and Cognitive Mechanisms of Learning and Creativity}
\begin{itemize}[topsep=-2pt,left=12pt,itemsep=0pt]
    \item \textit{Abbreviated as \textbf{\textcolor{myblue}{Neural Mechanisms}}}
    \item This cluster focuses on the brain and cognitive systems that support learning, memory, and creative thinking. 
    Research covers neuroimaging (e.g., \citealp{bookheimer2002functional,kanwisher1997fusiform}), computational modeling (e.g., \citealp{anderson2013adaptive,tenenbaum2006theory}), and the dynamic interplay between neural circuits and cognitive function (e.g., \citealp{baddeley2003working,wang2018prefrontal}). 
\end{itemize}

\vspace{0.5em}
\textbf{6) Psychometrics, and Judgment and Decision-Making}
\begin{itemize}[topsep=-2pt,left=12pt,itemsep=0pt]
    \item \textit{Abbreviated as \textbf{\textcolor{mypurple}{Psychometrics \& JDM}}}
    \item This cluster includes the study of psychometric measurement and the study of human decision processes.
    It includes scale development (e.g., \citealp{hamilton2016development,john1999bigfive}), modeling of cognitive biases (e.g., \citealp{nickerson1998confirmation,tversky1974judgment}), and understanding how people assess risk (e.g., \citealp{lejuez2002evaluation,mishra2011individual}), probability (e.g., \citealp{bar1980base,cosmides1996humans}), and outcomes (e.g., \citealp{hornsby2020decisions,oliver1994outcome}). 
\end{itemize}

For the sake of brevity, these clusters will hereinafter be referred to by their respective abbreviations.
\section{Results}\label{sec:5}

\subsection{How is psychology research integrated into LLM research?}\label{sec:5.1}

\begin{wrapfigure}{l}{0.49\linewidth}
    \centering
    \includegraphics[width=\linewidth]{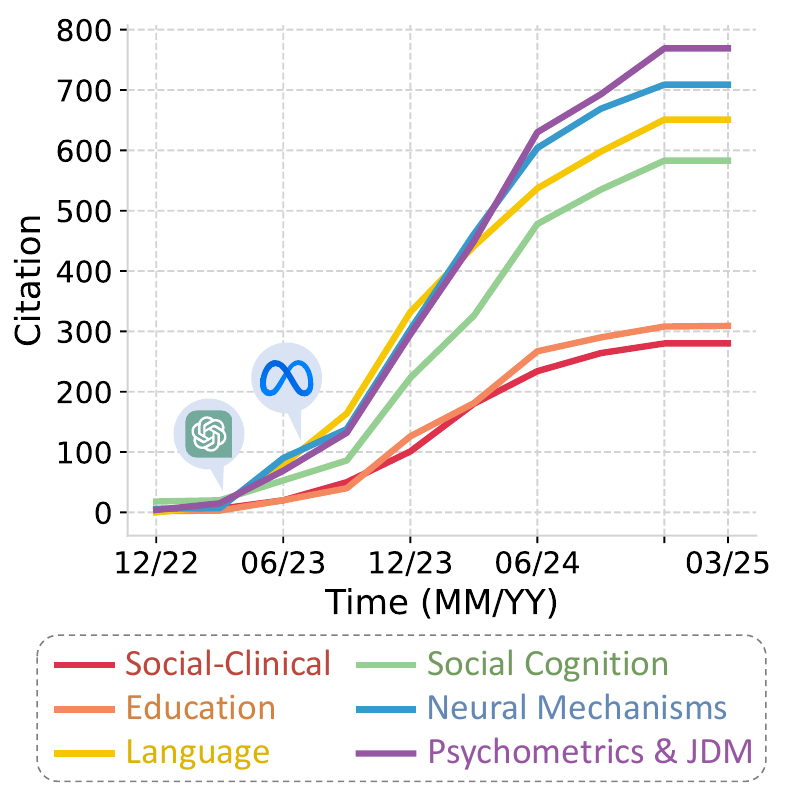}
    \caption{Overall Citation Flow from LLM Research Papers to Psychology Papers Over Time.}
    \label{fig:citation_all}
\end{wrapfigure}

\begin{figure}[h]
    \centering
    \includegraphics[width=\linewidth]{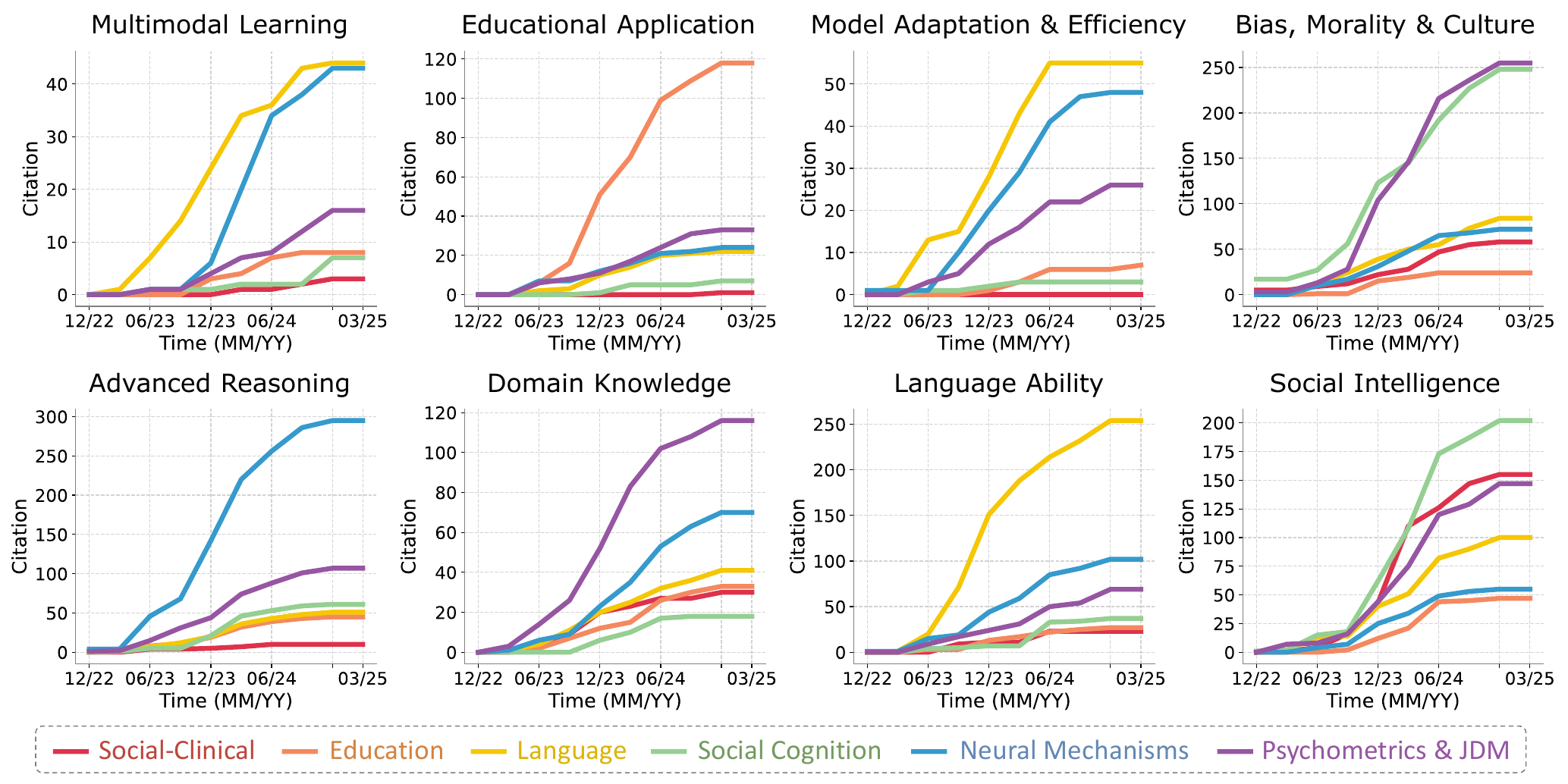}
    \caption{Citation Flow from LLM Research Papers to Psychology Papers Over Time. \\
\textit{Note:} Each subfigure presents a grouping of research papers on LLMs organized by cluster, with colors indicating the corresponding clusters in the psychology literature. Abbreviated cluster labels are displayed for brevity; complete cluster names are provided in §\ref{sec:2}.}
    \label{fig:citation_split}
\end{figure}

Once the citation analysis was finished, we were able to clearly observe the interrelationships between the eight LLM research clusters and the six psychology research clusters. 
The overall analysis results are presented in Fig.~\ref{fig:bipart}. Fig.~\ref{fig:citation_all} illustrates the temporal citation trends, while Fig.~\ref{fig:citation_split} provides a more detailed view of how each LLM research cluster has cited papers from the psychology clusters across different time periods.
We observe the following patterns through our analysis.

\vspace{5pt}
\textbf{Finding 1: LLM research has increasingly cited psychology research in recent years.}

We first observed a growing trend in the incorporation of psychology research within the LLM literature over time.
As shown in Fig.~\ref{fig:citation_all}, the LLM research community has increasingly emphasized insights from psychology.
This trend began around March 2023, when researchers started citing certain clusters of psychology research — notably the \textit{Neural Mechanism}, \textit{Language}, and \textit{Psychometrics \& JDM} clusters, which were among the earliest to receive attention.
Subsequently, around July 2023, the volume of psychology-related citations in LLM research saw a marked increase.
Later, by mid to late 2024, the overall growth in citation volume began to slow down.

We speculate that this emerging trend can be understood in several ways.
First, the initial references to psychology in LLM research around March 2023 appear to coincide with the release of GPT-3.5-Turbo and GPT-4.
This event may have sparked heightened academic interest in the inner workings of LLMs.
At this early stage, researchers began drawing upon clusters most closely related to LLM mechanisms (\textit{Neural Mechanism} and \textit{Language}), as well as the cluster most relevant to model evaluation (\textit{Psychometrics \& JDM}).
Around July 2023, the noticeable uptick in psychology-related citations may be linked to the release of open-source models like Llama2.
The accessibility and flexibility of open-source models likely facilitated interdisciplinary collaboration, allowing researchers to more freely modify model architectures and behaviors to explore psychologically informed hypotheses and experiments.
By the latter half of 2024, although citations of psychology in LLM research continued to rise, the rate of increase appeared to slow. 
This stabilization may suggest that core psychology research has largely been assimilated into the LLM research framework.

\vspace{5pt}
\textbf{Finding 2: Psychology is broadly integrated into LLM research, with particular emphasis on certain clusters (i.e., \textit{Psychometrics \& JDM} and \textit{Neural Mechanisms}).}

We also found that psychology research is extensively referenced in LLM studies, with nearly all eight LLM research clusters drawing from multiple psychology clusters rather than a single domain (except for \textit{Model Adaptation \& Efficiency} $\leftarrow$ \textit{Social-Clinical}; see Fig.~\ref{fig:bipart}).
This suggests that LLM research engages with various areas of psychology, depending on the specific research questions or methodological needs.

Moreover, certain psychology research clusters are cited more frequently.
As shown in Fig.~\ref{fig:citation_all}, the distribution of citation frequency reveals a distinct hierarchical pattern: \textit{Psychometrics \& JDM} $\approx$ \textit{Neural Mechanism} $>$ \textit{Language} $>$ \textit{Social Cognition} $\gg$ \textit{Education} $\approx$ \textit{Social-Clinical}. 
This likely reflects differences in methodological or conceptual alignment, with some research clusters being more closely aligned with the core objectives of current LLM research than others.

For example, the \textit{Psychometrics \& JDM} cluster contributes important tools for modeling and evaluating cognitive-like behaviors in LLMs. 
Foundations such as Classical Test Theory~\citep[CTT;][]{NOVICK19661} and Item Response Theory~\citep[IRT;][]{lord1980applications} inform assessment frameworks, while work in judgment and decision-making (JDM) offers analogies for understanding LLM reasoning and uncertainty(e.g.,~\citealp{ALABED2022121786, Placani2024}). 
Similarly, the prominence of the \textit{Neural Mechanism} cluster underscores the foundational role of neuroscience and cognitive psychology in shaping LLM research. 
Seminal contributions from these fields, such as Hebbian learning \citep{hebb1949organization} and early connectionist models like the perceptron \citep{rosenblatt1958perceptron}, have directly influenced the design of neural architectures, including modern deep learning models like transformers \citep{NEURIPS2022_8c22e5e9,kaplan2020scaling}.

In contrast, \textit{Education} and \textit{Social-Clinical} clusters are cited less frequently, which we speculate may be due to several reasons.
First, research in these clusters often relies on long-term, large-scale human feedback collection, which can slow the pace of LLM-related advancements. 
Second, studies in these clusters often involve sensitive data, such as student or patient information, which raises significant privacy concerns and makes data sharing and reuse more difficult due to strict ethical and legal constraints.
Third, many contributions from these clusters are commonly submitted to the Human-Computer Interaction (HCI) community ~\citep{BLANDFORD201941,10.1145/3658619.3658621,10.1145/3290605.3300475}, which is not included in this survey due to methodological differences.

\vspace{5pt}
\textbf{Finding 3: Different clusters of LLM research exhibit different tendencies in referencing psychology research.}

After examining the overall patterns of how LLM research has cited psychology literature, we further explored the specific clusters within LLM research (see Fig.~\ref{fig:citation_split}).
Overall, different clusters of LLM research tend to favor different clusters of psychology research, reflecting variations in research focus.

For example, \textit{Educational Application} shows a clear citation preference for \textit{Education}, while \textit{Advanced Reasoning} tends to favor citations from \textit{Neural Mechanisms}. 
This pattern may be explained by the strong conceptual alignment between the LLM research cluster and the Psychology research cluster. 
Specifically, educational applications naturally draw on foundational work in educational psychology; whereas reasoning tasks tend to rely on insights from cognitive neuroscience to model complex inferential behavior, which can be traced back to neurons in Artificial Neural Networks~\citep[ANN;][]{485891}.

Other clusters, such as \textit{Model Adaptation \& Efficiency} and \textit{Social Intelligence} draw upon a substantially broader range of psychology clusters. 
This likely reflects the greater conceptual complexity inherent in constructs such as adaptation and awareness, which place higher demands on researchers to cite multiple aspects of psychology research.
For example, \textit{Social Intelligence} requires modeling human mental states and traits such as emotions~\citep{ekman1992argument}, cultural beliefs~\citep{doi:10.1073/pnas.0903616106}, mental health~\citep{elliott2018empathy}, and personality~\citep{john1999bigfive}.
This drives frequent citation of work from the \textit{Social Cognition} and \textit{Social-Clinical} psychology research clusters.
At the same time, evaluating social awareness often involves extensive human-subject studies, which frequently result in citations of inter-rater reliability measures~\citep{cohen1960coefficient, fleiss1971measuring, 10.1093/ije/dyq191} from the \textit{Psychometrics \& JDM} cluster.

\subsection{Which psychology theories/frameworks are most commonly used, and which remain underexplored in LLM research?}\label{sec:5.2}

Building upon the analysis presented in §\ref{sec:5.1}, which examined the overall citation patterns of psychology research within the LLM literature, we now undertake a more granular investigation into how LLM research engages with specific psychology theories and frameworks within each identified psychology research cluster. 
For each cluster, we highlight both the most frequently cited theories/frameworks and those that remain underutilized or overlooked.

In addition to the cluster-specific analysis, we have identified the most influential psychology theories and frameworks across the entire LLM research landscape.
By examining the top 10 most frequently cited psychology papers across all clusters (see Table~\ref{tab:top_10}), we find that the majority belong to the \textit{Psychometrics \& JDM} and \textit{Neural Mechanisms} clusters, further supporting Finding 2 discussed in §\ref{sec:5.1}. 
The key theories represented in these influential works are Classical Test Theory (CTT) and Theory of Mind (ToM).

\begin{table}
    \small
    \centering
    \begin{tabular}{lll}
    \toprule
    \textbf{Topic} & \textbf{Paper} & \textbf{Related Theory/Framework} \\
    \midrule
    \multirow{2}{*}{\textcolor{mypurple}{Psychometrics \& JDM}} & Measuring Nominal Scale Agreement & \multirow{2}{*}{Classical Test Theory} \\
    & Among Many Raters~\citep{fleiss1971measuring} & \\
    \midrule
    \multirow{2}{*}{\textcolor{mypurple}{Psychometrics \& JDM}} &A Coefficient of Agreement for Nominal & \multirow{2}{*}{Classical Test Theory} \\
    & Scales~\citep{cohen1960coefficient} & \\
    \midrule
    \multirow{2}{*}{\textcolor{myblue}{Neural Mechanisms}}&Does the Chimpanzee Have A Theory of & \multirow{2}{*}{Theory of Mind} \\
    & Mind? \citep{Premack1978-PREDTC-3} & \\
    \midrule
    \multirow{2}{*}{\textcolor{mypurple}{Psychometrics \& JDM}} &The Proof and Measurement of Association & \multirow{2}{*}{Classical Test Theory} \\
    & between Two Things~\citep{10.1093/ije/dyq191} & \\
    \midrule
    \multirow{2}{*}{\textcolor{myblue}{Neural Mechanisms}} &Catastrophic Forgetting in Connectionist & \multirow{2}{*}{Complementary Learning Systems} \\
    & Networks~\citep{french1999catastrophic} & \\
    \midrule
    \multirow{2}{*}{\textcolor{myblue}{Neural Mechanisms}} &Does the Autistic Child Have A “Theory  & \multirow{2}{*}{Theory of Mind} \\
    & of Mind”?~\citep{BARONCOHEN198537} & \\
    \midrule
    \multirow{2}{*}{\textcolor{mypurple}{Psychometrics \& JDM}} &A Technique for the Measurement of & \multirow{2}{*}{Likert Scale} \\
     & Attitudes~\citep{likert1932technique} & \\
    \midrule
    \multirow{2}{*}{\textcolor{mypurple}{Psychometrics \& JDM}} &Judgment under Uncertainty: Heuristics and & \multirow{2}{*}{Heuristics and Biases Program}\\
    & Biases~\citep{doi:10.1126/science.185.4157.1124} & \\
    \midrule
    \multirow{4}{*}{\textcolor{mygreen}{Social Cognition}} & Beliefs about Beliefs: Representation and & \multirow{4}{*}{Theory of Mind} \\
    & Constraining Function of Wrong Beliefs in & \\
    & Young Children's Understanding of & \\
    & Deception~\citep{WIMMER1983103} & \\
    \midrule
    \multirow{2}{*}{\textcolor{myorange}{Education}} & A Revision of Bloom's Taxonomy: An & \multirow{2}{*}{Bloom's Taxonomy} \\
    & Overview~\citep{Krathwohl01112002} & \\
    \bottomrule
    \end{tabular}
    \caption{Top 10 Most Cited Psychology Papers in LLM Research.}
    \label{tab:top_10}
\end{table}

\begin{table}
    \small
    \centering
    \begin{tabular}{ll}
    \toprule
    \textbf{Sub-Topic} & \textbf{Theory/Framework} \\
    \midrule
    \multirow{3}{*}{\begin{tabular}{l}
        Analysis and Application of Health Communication
    \end{tabular}} & \cellcolor{myred!70}Cognitive Behavioral Therapy $\blacktriangleleft$ \\
    & \cellcolor{myred!18}The Belmont Report (Ethical Principals) \\
    & \cellcolor{myred!18}Motivational Interviewing \\
    \midrule
    \multirow{4}{*}{\begin{tabular}{l}
        Assessment Tools and Diagnostic Frameworks in Health
    \end{tabular}} & \cellcolor{myred!45}The Diagnostic and Statistical Manual \\
    & \cellcolor{myred!45}of Mental Disorders $\blacktriangleleft$\\
    & \cellcolor{myred!15}Five Factor Model \\
    & \cellcolor{myred!5}The Dark Triad \\
    \midrule
    \multirow{3}{*}{\begin{tabular}{l}
        Therapeutic Processes, Intervention Methods, and the \\
        Therapeutic Relationship
    \end{tabular}} & \cellcolor{myred!70}Cognitive Behavioral Therapy $\blacktriangleleft$ \\
    & \cellcolor{myred!36}The Working Alliance \\
    & \cellcolor{myred!18}Motivational Interviewing \\
    \midrule
    \multirow{3}{*}{\begin{tabular}{l}
        Stigma, Discrimination, and Health Disparities
    \end{tabular}} & \cellcolor{myred!47}Goffman's Theory of Stigma $\blacktriangleleft$\\
    & \cellcolor{myred!37}Minority Stress Model \\
    & \cellcolor{myred!22}Intersectionality \\
    \bottomrule
    \end{tabular}
    \caption{Subtopics and Corresponding Top Theories or Frameworks in the \textit{Social-Clinical} Cluster. \\ \textit{Note:} Cell opacity represents citation frequency; black triangles indicate the three most frequently cited theories/frameworks.}
    \label{tab:sub_0}
\end{table}

\begin{figure}
    \centering
    \includegraphics[width=\linewidth]{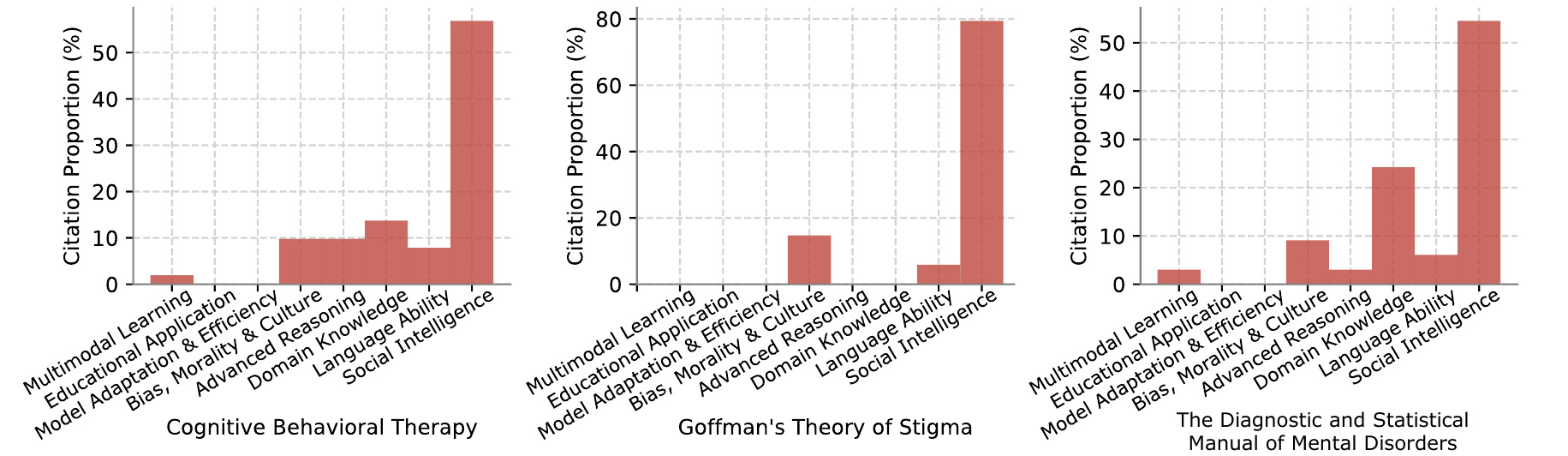}
    \caption{Citation Distribution for the Top Three Theories/Frameworks in the \textit{Social-Clinical} Cluster Across Eight LLM Research Topics. }
    \label{fig:theory_0}
\end{figure}

\subsubsection{Social-clinical psychology of mental health and intervention}\label{sec:5.2.1}

\paragraph{Popular theories/frameworks}
In the \textit{Social-Clinical} cluster, the three most frequently referenced psychology theories/frameworks in the surveyed LLM research papers are \textit{Cognitive Behavioral Therapy} (CBT; e.g.,~\citealp{beck2011cbt}), \textit{Goffman’s Theory of Stigma} (GTS; e.g.,~\citealp{10.1093/sf/43.1.127}), and \textit{the Diagnostic and Statistical Manual of Mental Disorders} (DSM; e.g.,~\citealp{dsm5}). 
Their citation distributions across the eight LLM research topics are shown in Fig.~\ref{fig:theory_0}.

\textbf{Cognitive Behavioral Therapy (CBT)} is a psychotherapeutic framework that focuses on the interconnectedness of thoughts, emotions, and behaviors, aiming to help individuals identify and modify negative or maladaptive patterns.
It has been demonstrated to be effective for a range of problems, including alcohol and drug use problems, marital problems, and severe mental illness.
In this survey, CBT is the most frequently referenced theory/framework in the \textit{Social-Clinical} cluster.
LLM researchers primarily draw on paradigms from the CBT framework to develop models related to psychotherapy, resulting in 51 citations from the surveyed LLM research papers.
For example, \citet{wang-etal-2024-patient} used the CBT framework and LLMs to simulate virtual patients with various cognitive distortions, which could serve as a training tool for therapists to help them learn how to effectively formulate real cognitive models.
Similarly, \citet{xiao-etal-2024-healme} also adopted the CBT framework and proposed an LLM-based mental enhancement model (empathetic dialogue system) for cognitive framing therapy.
LLM research has also explored integrating LLMs into various stages of the CBT process.
For example, \citet{lissak-etal-2024-colorful} examined how LLMs could offer emotional support to queer adolescents, and \citet{gabriel-etal-2024-ai} evaluated the feasibility and ethical considerations of applying LLMs in mental health support.

\textbf{Goffman’s Theory of Stigma (GTS)} is a theory that explores how individuals with attributes deemed undesirable by society experience social disapproval, exclusion, and discrimination. 
It emphasizes the role of societal norms and interactions in labeling individuals as deviant, leading to a spoiled social identity and altered self-concept.
GTS has been influential in understanding the social dynamics of mental illness, physical disability, addiction, and other marginalized statuses, highlighting how stigma can affect access to resources, treatment engagement, and psychological well-being.
LLM researchers have primarily drawn on GTS to explore whether LLMs exhibit bias and discrimination, and whether they amplify existing stigmas, resulting in 34 citations from the surveyed LLM research papers.
For example, \citet{an-etal-2024-large} draws on the GTS that conceptualizes names as identity cues that function as social labels. 
By using gendered and ethnically marked names, they examine whether LLMs implicitly activate stereotypical associations tied to specific social groups.
Similarly, \citet{morabito-etal-2024-stop} adopts the GTS point of stigma not as a discrete or isolated event, but as a structural and dynamic process.
On this basis, this paper designs a dataset consisting of progressively intensified offensive language to model the escalation of stigmatization.

\textbf{Diagnostic and Statistical Manual of Mental Disorders (DSM)} is a standardized classification framework developed by the American Psychiatric Association for diagnosing mental health conditions.
It provides clinicians with a common language and specific diagnostic criteria based on observable symptoms and clinical features.
The DSM is widely used in research, clinical practice, and insurance reporting, and plays a central role in shaping the understanding, treatment, and categorization of mental disorders across diverse populations and settings.
LLM researchers have primarily drawn on the DSM framework to guide the application of LLMs in the mental health domain, resulting in 33 citations from the surveyed LLM research papers.
The DSM provides clinical guidance, standardized symptom definitions, diagnostic labels, and decision-making criteria, thereby enhancing the scientific rigor, accuracy, and interpretability of LLM-based approaches.
For example, \citet{rosenman-etal-2024-llm} leverages the DSM framework to enable LLMs to interpret unstructured psychological interviews for more accurate automated mental health assessments.
Similarly, \citet{kang-etal-2024-cure}, building on the DSM framework, integrates contextual information about symptoms to design a novel approach for LLM-based psychiatric disorder detection, aiming to reduce potential errors in automated symptom recognition.

\paragraph{Under-explored theories/frameworks}
In addition to the three theories/frameworks widely adopted by most LLM researchers, we also list three others that are closely related to \textit{Social-Clinical} cluster but have received relatively little attention in current LLM studies.
These theories have been extensively applied and have had a significant impact in the field of psychology. 
They also hold the potential to offer new perspectives and valuable insights for LLM research, making them well worth further exploration and consideration.

\textbf{Biopsychosocial Model (BM)} is a holistic framework that conceptualizes health and illness as the result of an interaction between biological, psychological, and social factors. 
It recognizes that mental and physical health are influenced not only by genetic or physiological processes, but also by emotions, thoughts, behaviors, relationships, and environmental contexts.
BM is widely used in clinical assessment and treatment planning, especially in fields such as psychiatry, chronic pain management, and behavioral medicine, promoting a more integrated and person-centered approach to care.
In LLM research, the BM can serve as a guiding framework for modeling user behavior and tailoring responses in a human-centered manner.
By considering users’ emotional states, cognitive patterns, and social contexts, LLMs can generate responses that are more empathetic, contextually relevant, and effective in addressing users’ complex needs.
This approach enhances user trust, satisfaction, and long-term engagement by aligning model behavior with the multifaceted nature of human experience.

\textbf{Critical Race Theory (CRT)} is a framework that examines how race and racism are embedded within social structures, institutions, and policies. 
It challenges the notion of racial neutrality and emphasizes that systemic inequality is maintained through laws, cultural narratives, and power dynamics that privilege dominant groups.
CRT has been applied across disciplines such as education, public health, and sociology to highlight the lived experiences of marginalized communities and to advocate for structural change, equity, and social justice.
In LLM research, CRT can serve as a lens to critically assess and mitigate biases in model outputs, training data, and deployment contexts.
By incorporating CRT principles, researchers can better identify how LLMs might perpetuate racial stereotypes or inequities, and develop strategies to promote fairness, inclusivity, and accountability.
This includes refining datasets, adopting more equitable evaluation metrics, and designing interaction protocols that center the voices and perspectives of historically marginalized users.

\textbf{Health Belief Model (HBM)} is a framework that explains health-related behaviors by focusing on individuals’ perceptions of risk and benefits. 
It posits that behavior change is influenced by key factors such as perceived susceptibility to a health issue, perceived severity of the condition, perceived benefits of taking action, and perceived barriers to action.
HBM has been widely applied in public health to design interventions that promote preventive health behaviors, such as vaccination, screening, and lifestyle modification, by addressing motivational and cognitive determinants of decision-making.
In LLM research, the HBM can also serve as a theoretical framework for understanding and guiding user behavior. 
Researchers can leverage key components of the HBM, such as perceived susceptibility, perceived severity, perceived benefits, and perceived barriers, to design more persuasive and personalized dialogue strategies, thereby enhancing interaction quality and the model’s ability to influence user behavior. 
By identifying users’ motivations and concerns when responding to model-generated suggestions, LLMs can dynamically adjust their outputs to improve the adoption rate and trustworthiness of their recommendations.

\begin{table}
    \small
    \centering
    \begin{tabular}{ll}
    \toprule
    \textbf{Sub-Topic} & \textbf{Theory/Framework} \\
    \midrule
    \multirow{3}{*}{\begin{tabular}{l}
        Cognitive, Social, and Developmental
        Factors of Academic \\
        Achievement \\
    \end{tabular}} & \cellcolor{myorange!14}Dweck’s Mindset Theory \\
    & \cellcolor{myorange!10}The Matthew Effect in Education \\
    & \cellcolor{myorange!10}Five Factor Model \\
    \midrule
    \multirow{3}{*}{\begin{tabular}{l}
        Reading Comprehension and Vocabulary Development 
    \end{tabular}} & \cellcolor{myorange!55}Instructional Scaffolding $\blacktriangleleft$ \\
    & \cellcolor{myorange!17}Text Structure Theory \\
    & \cellcolor{myorange!9}Keyword Mnemonic Method \\
    \midrule
    \multirow{3}{*}{\begin{tabular}{l}
        Pedagogy, Cognitive Processes, and
        Communication Strategies 
    \end{tabular}} & \cellcolor{myorange!70}Self-Regulated Learning $\blacktriangleleft$\\
    & \cellcolor{myorange!34}Bandura's Social Cognitive Theory \\
    & \cellcolor{myorange!14}Dweck’s Mindset Theory \\
    \midrule
    \multirow{3}{*}{\begin{tabular}{l}
        Cognitive Science of Learning and
        Instructional Design 
    \end{tabular}} & \cellcolor{myorange!35}Bloom’s Taxonomy $\blacktriangleleft$\\
    & \cellcolor{myorange!25}Cognitive Load Theory \\
    & \cellcolor{myorange!5}Gagné's Conditions of Learning Theory \\
    \bottomrule
    \end{tabular}
    \caption{Subtopics and Corresponding Top Theories or Frameworks in the \textit{Education} Cluster. \\ \textit{Note:} Cell opacity represents citation frequency; black triangles indicate the three most frequently cited theories/frameworks.}
    \label{tab:sub_1}
\end{table}

\begin{figure}
    \centering
    \includegraphics[width=\linewidth]{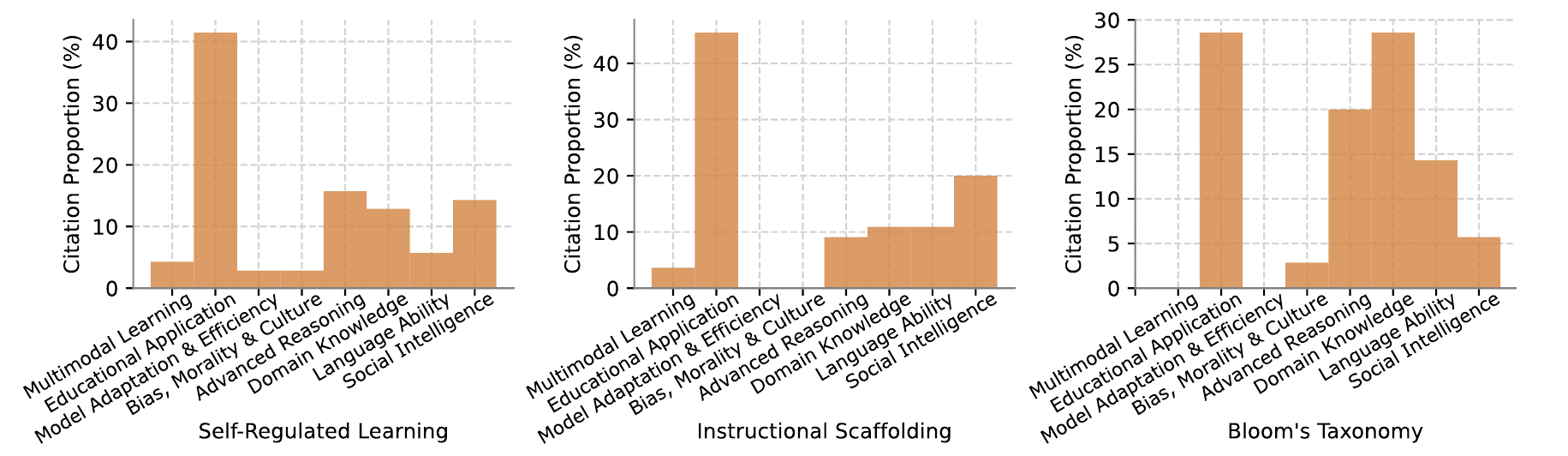}
    \caption{Citation distribution for the top three theories/frameworks in the \textit{Education} cluster across eight LLM research topics.}
    \label{fig:theory_1}
\end{figure}

\subsubsection{Learning, teaching design, and educational development}\label{sec:5.2.2}

\paragraph{Popular theories/frameworks}
In the \textit{Education} cluster, the three most frequently referenced psychology theories/frameworks in the surveyed LLM research papers are \textit{Self-Regulated Learning} (SRL; e.g.,~\citealp{graham1993srl}), \textit{Instructional Scaffolding} (IS; e.g.,~\citealp{penny1991moving}), and \textit{Bloom's Taxonomy} (BT; e.g.,~\citealp{bloom1956taxonomy}).
Their citation distributions across the eight LLM research topics are shown in Fig.~\ref{fig:theory_1}.

\textbf{Self-Regulated Learning (SRL)} is a theory that highlights the active role learners play in their own educational processes by setting goals, monitoring progress, and reflecting on outcomes. 
It emphasizes the cyclical interaction between cognitive, metacognitive, motivational, and behavioral components, enabling learners to strategically manage their learning environments and efforts. 
SRL has been shown to enhance academic performance, foster lifelong learning skills, and support students across diverse contexts, including online learning, special education, and higher education settings.
In this survey, SRL is the most frequently referenced theory/framework in the \textit{Education} cluster.
LLM researchers have drawn on the SRL to inspire the design of self-improving LLM systems, resulting in 70 citations from the surveyed LLM research papers.
The principles of SRL, especially self-monitoring and iterative feedback, have guided work in enabling LLMs to autonomously refine their outputs across domains, including problem solving \citep{gou2024critic,wang-etal-2024-bridging}, fact-checking \citep{ying-etal-2024-llms,yu2024kola}, code generation \citep{huang2024effilearner}, and data synthesis \citep{shi-etal-2024-culturebank}, thereby leveraging their reasoning capacity and internal knowledge structures.
LLM also adopted the SRL theory to support education applications.
For example, \citet{macina-etal-2023-mathdial} involved LLMs in the construction of dialog-based tutoring datasets to support reflective math learning, and \citet{borges-etal-2024-teach} explored how LLM-generated feedback can model and enhance SRL strategies in educational contexts.

\textbf{Instructional Scaffolding (IS)} is an educational framework that emphasizes the gradual support of learners as they develop new skills and understanding, with the ultimate goal of fostering independent competence. 
It involves the strategic use of guidance, prompts, and resources by instructors to bridge the gap between what learners can do alone and what they can achieve with assistance. 
IS has been shown to be effective in various learning contexts, including literacy development, problem-solving, and complex conceptual learning across age groups and subject areas.
In LLM research, IS has inspired the design of prompting strategies that position LLMs as instructional agents capable of guiding users through incremental reasoning. It received 55 citations from the surveyed LLM research papers.
For example, \citet{daheim-etal-2024-stepwise}, \citet{sonkar-etal-2024-pedagogical}, and \citet{wang-etal-2024-unleashing} draw on IS principles to craft pedagogically informed prompts, enabling LLMs to lead users or other models through intermediate steps toward task completion.
On the other hand, IS also offers a valuable conceptual lens for treating LLMs as adaptive learners that benefit from guided feedback.
In this paradigm, LLMs mirror students who improve through iterative supervision, corrections, and example-based guidance.
Such applications enhance the model’s ability to internalize human preferences and refine performance over time.
For example, \citet{tian-etal-2024-theory} and \citet{wang-etal-2024-bridging} implement feedback-driven training and refinement loops that emulate scaffolded learning processes, allowing LLMs to self-correct and align more closely with desired outcomes.

\textbf{Bloom’s Taxonomy (BT)} is a hierarchical educational theory that categorizes cognitive skills into progressive levels, aiming to promote deeper learning and critical thinking.
It serves as a foundational model for curriculum development, instructional design, and assessment strategies across educational settings, guiding learners from basic knowledge recall to advanced analytical and creative thinking.
BT has been adopted in LLM research to guide the design and evaluation of model capabilities, and it was cited 35 times in the surveyed LLM research papers.
By providing a structured hierarchy of cognitive complexity, BT enhances the interpretability, rigor, and educational alignment of LLM-based benchmarks and evaluation methodologies.
For example, \citet{10.5555/3737916.3739523}, \citet{cao-etal-2024-structeval}, \citet{wang-etal-2023-newton}, and \citet{yu2024kola} reflect BT principles in benchmark construction, while \citet{ying2024automating} incorporates BT into dynamic benchmarking strategies.
On the other hand, BT also serves as a conceptual lens for interpreting the internal structure of LLMs.
For example, \citet{10.5555/3737916.3742165} explores how BT can shed light on the internal cognitive process of LLMs, and \citet{wang-etal-2024-knowledge-mechanisms} draws on BT to investigate the underlying mechanisms of knowledge representation in language models.

\paragraph{Under-explored theories/frameworks}

We also list three other theories/frameworks that are closely related to the \textit{Education} cluster but have received relatively little attention in current LLM studies.

\textbf{Bronfenbrenner's Ecological Systems Theory (BEST)} is a developmental theory that emphasizes the multiple layers of environmental influence on an individual’s growth and behavior. 
It outlines how individuals are embedded within a series of interrelated systems, ranging from immediate settings like family and school (microsystem) to broader societal and cultural forces (macrosystem). 
The theory highlights the dynamic interactions between these systems and how changes in one layer can ripple through others, shaping developmental outcomes over time. 
BEST is widely applied in fields such as education, psychology, and public health to understand and support human development within context.
In LLM research, BEST can serve as a framework for contextualizing user interactions by considering the multilayered influences on user behavior and preferences. 
By modeling users within their broader ecological environments, such as cultural norms, social relationships, and institutional contexts, LLMs can tailor their responses to better align with users’ lived experiences, thereby enhancing relevance, empathy, and user engagement.

\textbf{Simple View of Reading (SVR)} is a theory that posits reading comprehension as the product of two primary components: decoding and linguistic comprehension. 
According to SVR, proficient reading occurs when individuals can accurately translate written symbols into spoken language (decoding) and effectively understand spoken language (comprehension).
This model has been widely supported in empirical research and is used to inform assessments and interventions for reading difficulties, such as dyslexia and language impairment.
In LLM research, SVR can serve as a framework for evaluating and enhancing models' reading comprehension abilities. 
By separately analyzing a model's decoding-like abilities (e.g., text recognition and parsing) and its linguistic comprehension abilities (e.g., understanding semantics and context), researchers can better identify specific strengths and limitations. 
This dual-component perspective can also guide the development of more targeted training strategies, improving both surface-level processing and deep understanding in LLM outputs.

\textbf{Self-Determination Theory (SDT)} is a theory that focuses on human motivation, emphasizing the role of innate psychological needs, autonomy, competence, and relatedness, in fostering self-motivated and healthy behavior.
It has been widely applied across various domains, including education, healthcare, workplace settings, and psychotherapy, demonstrating effectiveness in enhancing well-being, intrinsic motivation, and sustained behavior change.
In LLM research, SDT can serve as a guiding framework to promote user engagement and satisfaction by aligning model outputs with users’ psychological needs. 
For example, LLMs can support autonomy by offering choices or allowing users to guide the direction of interactions, enhance competence by providing clear, constructive feedback, and foster relatedness through empathetic and personalized responses. 
By integrating SDT principles, LLMs can improve not only the effectiveness of user interactions but also long-term trust and continued use.

\begin{table}
    \small
    \centering
    \begin{tabular}{ll}
    \toprule
    \textbf{Sub-Topic} & \textbf{Theory/Framework} \\
    \midrule
    \multirow{3}{*}{\begin{tabular}{l}
        Narrative, Discourse and Meaning-Making
    \end{tabular}} & \cellcolor{myyellow!40}Schema Theory $\blacktriangleleft$\\
    & \cellcolor{myyellow!12}Conceptual Metaphor Theory \\
    & \cellcolor{myyellow!4}Reader-Response Theory \\
    \midrule
    \multirow{3}{*}{\begin{tabular}{l}
        Phonetics, Prosody, and Interaction in Spoken \\
        Communication
    \end{tabular}} & \cellcolor{myyellow!26}Embodied Cognition Theory\\
    & \cellcolor{myyellow!23}Conversation Analysis \\
    & \cellcolor{myyellow!3}Articulatory Phonology \\
    \midrule
    \multirow{3}{*}{\begin{tabular}{l}
        Sociolinguistics, Culture, and Cross-Cultural \\
        Communication
    \end{tabular}} & \cellcolor{myyellow!39}Sapir-Whorf Hypothesis \\
    & \cellcolor{myyellow!12}The Emergence Theory of Language \\
    & \cellcolor{myyellow!6}Brown and Levinson’s Politeness Theory \\
    \midrule
    \multirow{3}{*}{\begin{tabular}{l}
        Pragmatic Inference and Information Processing \\
        in Dialogue
    \end{tabular}} & \cellcolor{myyellow!23}Grice’s Theory of Implicature \\
    & \cellcolor{myyellow!7}Rational Speech Act \\
    & \cellcolor{myyellow!6}Brown and Levinson’s Politeness Theory \\
    \midrule
    \multirow{3}{*}{\begin{tabular}{l}
        Cognitive and Neural Foundations of Language \\
        Processing
    \end{tabular}} & \cellcolor{myyellow!26}Embodied Cognition Theory \\
    & \cellcolor{myyellow!17}Construction-Integration Model \\
    & \cellcolor{myyellow!2}The Simple View of Reading \\
    \midrule
    \multirow{3}{*}{\begin{tabular}{l}
        Grammar, Lexicon, and Mental Representation
    \end{tabular}} & \cellcolor{myyellow!70}Connectionism vs. Symbolicism $\blacktriangleleft$ \\
    & \cellcolor{myyellow!47}Usage-Based Models of Language $\blacktriangleleft$ \\
    & \cellcolor{myyellow!25}Generative and Universal Grammar \\
    \midrule
    \multirow{3}{*}{\begin{tabular}{l}
        Computational Models of Language and Psycho- \\
        logical Processes
    \end{tabular}} & \cellcolor{myyellow!33}Gricean/Post-Gricean Pragmatics \\
    & \cellcolor{myyellow!18}Linguistic Inquiry and Word Count \\
    & \cellcolor{myyellow!12}Conceptual Metaphor Theory \\
    \bottomrule
    \end{tabular}
    \caption{Subtopics and Corresponding Top Theories or Frameworks in the \textit{Language} Cluster. \\ \textit{Note:} Cell opacity represents citation frequency; black triangles indicate the three most frequently cited theories/frameworks.}
    \label{tab:sub_2}
\end{table}

\begin{figure}
    \centering
    \includegraphics[width=\linewidth]{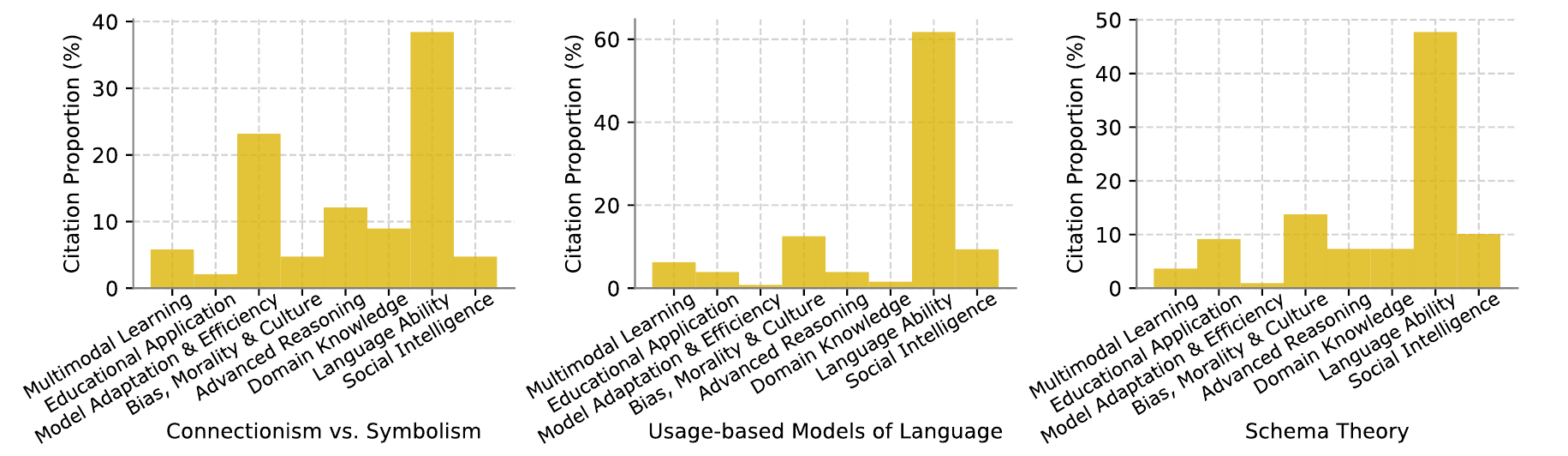}
    \caption{Citation distribution for the top three theories/frameworks in the \textit{Language} cluster across eight LLM research topics.}
    \label{fig:theory_2}
\end{figure}

\subsubsection{Language comprehension, pragmatic, and psycholinguistic}\label{sec:5.2.3}

\paragraph{Popular theories/frameworks}
In the \textit{Language} cluster, the three most frequently referenced psychology theories/frameworks in the surveyed LLM research papers are \textit{Connectionism and Symbolism}~\citep{faules1978sym,FODOR19883}, \textit{Usage-based Models of Language}~\citep{vonMengdenCoussé+2014+1+20}, and \textit{Schema Theory}~\citep{GRAESSER198259}.
Their citation distributions across the eight LLM research topics are shown in Fig.~\ref{fig:theory_2}.

\textbf{Connectionism and Symbolism} represent two contrasting approaches in cognitive science aimed at explaining human thought and information processing.
Connectionism relies on artificial neural networks to model mental processes, highlighting the role of distributed, parallel processing and learning from experience. Rather than explicit symbols, knowledge in connectionist models is stored in the patterns of connections and activation across the network.
In contrast, Symbolism, also known as the rule-based approach, posits that cognition operates through the manipulation of discrete symbols according to formal rules, akin to programming languages or logical systems. It emphasizes structured representations and explicit reasoning mechanisms.
While Connectionism offers strengths in pattern recognition, learning, and handling noisy or incomplete data, Symbolism excels at modeling rule-governed, high-level reasoning tasks.
Contemporary cognitive science often seeks integrative frameworks that draw upon the complementary strengths of both paradigms.

\textbf{Connectionism} serves as a foundational theory for understanding and designing neural architectures in LLMs.  
This theory directly informs research on the internal mechanics and adaptive capabilities of LLMs. 
For example, work on model compression and pruning, such as \citet{liu-etal-2024-llm} and \citet{ma2023llmpruner}, leverages the connectionist notion of representational redundancy to identify and remove unnecessary parameters while preserving functionality. 
Similarly, generalization studies like \citet{10.5555/3692070.3692585}, \citet{yang-etal-2024-unveiling}, and \citet{10.5555/3692070.3694452} explore how internal patterns learned through training allow LLMs to extrapolate to novel linguistic tasks and inputs, reflecting classic connectionist learning dynamics. 
Moreover, work on emergent linguistic capabilities—such as \citet{de-varda-marelli-2023-scaling} and \citet{10.5555/3618408.3619446}, illustrates how structured behaviors like syntax can arise from purely data-driven, neural processes, echoing one of Connectionism’s core claims that complex cognition need not rely on explicit symbolic rules.

\textbf{Symbolism}, by contrast, has significantly influenced LLM research in areas that require explicit reasoning, compositional understanding, and interoperability. 
For example, \citet{yuan-etal-2024-chatmusician} uses symbolic representations to enable structured musical instruction following, while \citet{tennenholtz2024demystifying} investigates how LLMs encode modular and interpretable meaning representations.
In formal linguistics, symbolic perspectives underpin works such as \citet{de-dios-flores-etal-2023-dependency}, \citet{minixhofer2023compoundpiece}, and \citet{nair-resnik-2023-words}, which align LLM behavior with traditional linguistic theories involving morphology, syntax, and dependency structures. 
Furthermore, in mathematical and logical reasoning, rule-based symbolic models can help LLMs handle problems requiring multi-step, deterministic computation, as demonstrated by \citet{imani-etal-2023-mathprompter} and \citet{zheng2024take}.

\textbf{Usage-based Models of Language (UBML)} are theories that emphasize the role of actual language use in shaping linguistic knowledge and structure.
Grounded in cognitive and functional approaches, UBML posits that language emerges from language users’ repeated experiences with specific linguistic forms in meaningful contexts.
These models highlight how frequency, context, and communicative function influence language acquisition, processing, and change.
UBML has been applied to a wide range of theoretical and applied domains in linguistics, including grammar development, lexical patterning, language variation, and second language acquisition.
LLM researchers have drawn on the UBML to understand and model how patterns of language usage influence LLM behavior and capabilities. 
For example, \citet{zhang-etal-2023-dont} applies UBML theory to explain how LLM performance across different languages is shaped by the frequency of usage in training data. 
Similarly, \citet{zeng-etal-2024-johnny} leverages the UBML perspective that language use shapes understanding to highlight LLMs’ sensitivity to communicative strategies. 
In their work, LLMs are treated as human-like persuaders, and real-world rhetorical techniques, such as emotional appeals, appeals to authority, and logical reasoning, are employed to achieve jailbreaks.
In addition, LLM researchers also use UBML theories as a basis for designing evaluation tasks aimed at assessing the LLMs' language abilities. 
For example, \citet{wachowiak-gromann-2023-gpt} draws on one of the core UBML theories, conceptual metaphor theory, to develop tasks that test whether LLMs can learn metaphorical mappings.

\textbf{Schema Theory} is a theory that posits that individuals organize knowledge into mental structures called schemas, which shape how they perceive, interpret, and respond to experiences. 
These schemas, developed through personal, social, and cultural narratives, guide meaning-making by filtering new information in accordance with existing beliefs and expectations.
Schema theory explains how people construct coherence in stories, comprehend language, and derive significance from interactions by drawing upon pre-existing cognitive and cultural templates. 
Schema theory has been applied extensively in education and language learning to explain how learners use prior knowledge to comprehend new information.
LLM researchers have drawn on schema theory to explore how LLMs emulate human-like reasoning and meaning construction through the activation of learned patterns.
For example, \citet{sui-etal-2024-confabulation} leverages the schema theory principle of generating coherent narratives based on pre-existing structures to reinterpret hallucinations in LLMs as a form of schema-like reasoning with potential narrative value.
Similarly, \citet{chen-etal-2024-generalizing} applies the schema activation mechanism to design a three-stage prompting framework (comprehend, associate, summarize) that simulates human cognitive processes to enhance generalization in multi-turn dialogue retrieval.
In addition, schema theory has also been used by LLM researchers as an evaluative framework to assess model abilities.
For example, \citet{wicke-wachowiak-2024-exploring} based on schema theory’s embodiment-oriented perspective to examine whether LLMs and VLMs can demonstrate human-like intuitions about spatial schemas (e.g., support, containment, path) in the absence of sensorimotor grounding.

\paragraph{Under-explored theories/frameworks}

We also list three other theories/frameworks that are closely related to the \textit{Language} cluster but have received relatively little attention in current LLM studies.

\textbf{Predictive Coding (PC)} is a theory that posits the brain as a predictive machine, constantly generating and updating models to anticipate sensory input.
It emphasizes the interplay between top-down predictions and bottom-up sensory signals, where discrepancies drive learning and perception.
PC has been influential in understanding perception, action, and cognition.
In LLM research, PC may inform the development of adaptive and context-aware models by treating dialogue as a dynamic process of prediction and error correction. 
By modeling user inputs as sensory signals and the model’s responses as top-down predictions, LLMs can iteratively refine their outputs based on user feedback and interaction history. 
This approach may enhance responsiveness, coherence, and personalization, enabling models to better align with user expectations and reduce communicative mismatches over time.

\textbf{Hofstede's Cultural Dimensions Theory (HCDT)} is a theory for understanding cultural differences through six key dimensions that influence how individuals perceive and interact with the world. 
It provides insights into national cultural values such as power distance, individualism vs. collectivism, masculinity vs. femininity, uncertainty avoidance, long-term orientation, and indulgence vs. restraint.
HCDT serves as a valuable tool for navigating cross-cultural communication, enabling individuals and organizations to recognize and adapt to cultural nuances that affect communication styles, conflict resolution, leadership expectations, and collaboration in international or multicultural settings.
In LLM research, HCDT can inform the development of culturally adaptive dialogue systems that are sensitive to users’ cultural backgrounds.
By incorporating insights from the six cultural dimensions, LLMs can tailor language, tone, and interaction strategies to align with users’ communication preferences and expectations.
This cultural alignment can enhance user engagement, reduce misunderstandings, and foster greater trust and effectiveness in diverse human-AI interactions.
Moreover, HCDT can serve as an evaluative framework to assess whether LLM-generated content appropriately reflects or adapts to cultural norms, providing a systematic way to analyze model performance across different cultural contexts.

\textbf{Grice's Cooperative Principle (GCP)} is a framework in pragmatics, explaining how effective and meaningful communication is achieved in conversation. 
GCP posits that speakers typically aim to be cooperative by contributing appropriately to the communicative context. 
This principle is supported by four conversational maxims (Quantity, Quality, Relation, and Manner), which guide interlocutors to provide the right amount of information, to be truthful, relevant, and clear. 
GCP helps interpret implied meanings, identify conversational implicatures, and understand communication breakdowns when the maxims are flouted or violated.
In LLM research, the GCP may serve as a foundational principle for designing more natural and contextually appropriate interactions. 
By aligning responses with the GCP framework, LLMs can enhance clarity, relevance, and trustworthiness in dialogue. 
Moreover, understanding when and how to strategically flout maxims (e.g., using understatement or irony) can help LLMs generate more nuanced and human-like communication.
GCP may also serve as an evaluation framework for measuring the pragmatic appropriateness of LLM responses. 
This involves assessing how well the model adheres to conversational maxims and whether it produces implicatures in a contextually coherent manner.

\begin{table}
    \small
    \centering
    \begin{tabular}{ll}
    \toprule
    \textbf{Sub-Topic} & \textbf{Theory/Framework} \\
    \midrule
    \multirow{3}{*}{\begin{tabular}{l}
        Collective Memory, Social Beliefs, and Self-Regulation
    \end{tabular}} & \cellcolor{mygreen!19}Bandura's Social Cognitive Theory \\
    & \cellcolor{mygreen!6}Inoculation Theory \\
    & \cellcolor{mygreen!3}Collective Memory Framework \\
    \midrule
    \multirow{3}{*}{\begin{tabular}{l}
        Emotion Pragmatics, Culture, and Cross-Cultural \\
        Communication
    \end{tabular}} & \cellcolor{mygreen!31}Expectancy Violations Theory \\
    & \cellcolor{mygreen!24}Empathy-Altruism Hypothesis \\
    & \cellcolor{mygreen!9}Emotions as Social Information Model \\
    \midrule
    \multirow{3}{*}{\begin{tabular}{l}
    	The Foundations, Judgment, and Development of Morality
    \end{tabular}} & \cellcolor{mygreen!35}Moral Foundations Theory \\
    & \cellcolor{mygreen!8}The "Big Three" of Morality \\
    & \cellcolor{mygreen!6}Kohlberg’s Stages of Moral Development \\
    \midrule
    \multirow{3}{*}{\begin{tabular}{l}
    	Persuasion, Deception, and Social Conflict
    \end{tabular}} & \cellcolor{mygreen!70}Theory of Mind $\blacktriangleleft$\\
    & \cellcolor{mygreen!60}Dual-Process Theory $\blacktriangleleft$\\
    & \cellcolor{mygreen!6}Inoculation Theory \\
    \midrule
    \multirow{3}{*}{\begin{tabular}{l}
    	Narrative, Empathy, and Psychological Influence
    \end{tabular}} & \cellcolor{mygreen!65}Simulation Theory $\blacktriangleleft$\\
    & \cellcolor{mygreen!13}Transportation Theory \\
    & \cellcolor{mygreen!7}Experience-Taking \\
    \midrule
    \multirow{3}{*}{\begin{tabular}{l}
    	Personality Traits and Social Behavior
    \end{tabular}} & \cellcolor{mygreen!19}Five Factor Model \\
    & \cellcolor{mygreen!9}The Dark Triad \\
    & \cellcolor{mygreen!3}HEXACO Model of Personality \\
    \midrule
    \multirow{3}{*}{\begin{tabular}{l}
    	Social Identity, Stereotypes, and Cultural Values
    \end{tabular}} & \cellcolor{mygreen!35}Moral Foundations Theory \\
    & \cellcolor{mygreen!27}Social Identity Theory \\
    & \cellcolor{mygreen!24}Stereotype Content Model \\
    \midrule
    \multirow{3}{*}{\begin{tabular}{l}
    	The Theory, Perception, and Social Function of Emotion
    \end{tabular}} & \cellcolor{mygreen!34}Basic Emotion Theory \\
    & \cellcolor{mygreen!25}Appraisal Theory of Emotion \\
    & \cellcolor{mygreen!17}Circumplex Model of Affect \\
    \bottomrule
    \end{tabular}
    \caption{Subtopics and Corresponding Top Theories or Frameworks in the \textit{Social Cognition} Cluster. \\ \textit{Note:} Cell opacity represents citation frequency; black triangles indicate the three most frequently cited theories/frameworks.}
    \label{tab:sub_3}
\end{table}

\subsubsection{Emotion, morality, and culture in social cognition}\label{sec:5.2.4}

\begin{figure}
    \centering
    \includegraphics[width=\linewidth]{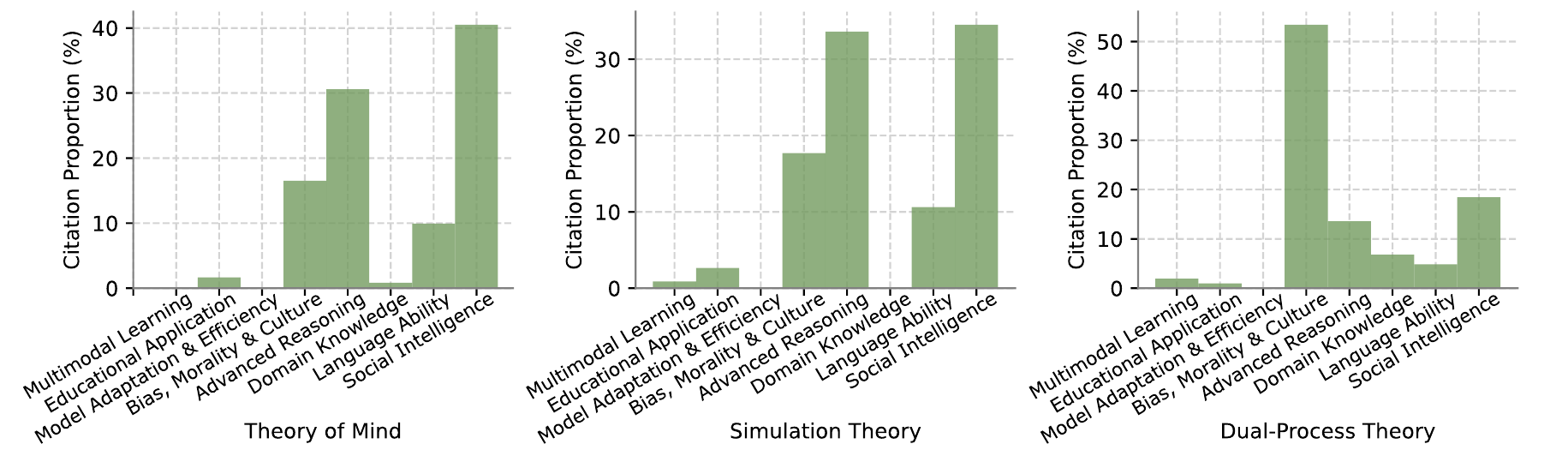}
    \caption{Citation distribution for the top three theories/frameworks in the \textit{Social Cognition} cluster across eight LLM research topics.}
    \label{fig:theory_3}
\end{figure}

\paragraph{Popular theories/frameworks}
In the \textit{Social Cognition} cluster, the three most frequently referenced psychology theories/frameworks in the surveyed LLM research papers are \textit{Theory of Mind}~\citep[ToM;][]{Premack1978-PREDTC-3,10.1093/biohorizons/hzq011}, \textit{Simulation Theory}~\citep{Shanton2010Simulation}, and \textit{Dual-Process Theory}~\citep{gawronski2013dpt}.
Their citation distributions across the eight LLM research topics are shown in Fig.~\ref{fig:theory_3}.

\textbf{Theory of Mind (ToM)} is a psychology theory of the ability to understand other individuals by ascribing mental states to them. In ToM, others' beliefs, desires, intentions, emotions, and thoughts are recognized as potentially different from one’s own. In other words, people use a theory of mind when analyzing, judging, and inferring others’ behaviors; a well-functioning ToM is crucial for success in everyday social interactions.
ToM is the most frequently referenced theory/framework in the \textit{Social Cognition} cluster, and it was cited 121 times in the surveyed LLM research papers.
In this cluster, LLM researchers have primarily drawn on ToM to interpret LLMs' social intelligence and their involvement in social interactions.
For example, various benchmarks (e.g.,~\citealp{kim-etal-2023-fantom,sabour-etal-2024-emobench}) and tests (e.g.,~\citealp{shapira-etal-2023-well}) were developed to examine LLMs' ToM capacity, which serves as a key indicator of their broader social intelligence. 
In addition, some research has focused on LLMs' use of ToM in specific contexts.
For example, \citet{zhao-etal-2024-large} evaluated LLM's ability to understand complex interpersonal relationships, 
and \citet{xu-etal-2024-opentom} incorporated personalized mental states into ToM assessment.
There have also been analyses on ToM for goal-oriented dialogues.
\citet{10.5555/3692070.3692228,chan-etal-2024-negotiationtom,lai-etal-2023-werewolf} explored how effectively LLMs can negotiate and persuade using ToM, and
\citet{wu-etal-2023-hi} introduced a new deception mechanism within higher-order ToM reasoning.

\textbf{Simulation Theory (ST)} is a theory of how people understand others by engaging in a form of empathetic simulation. It posits that humans anticipate and make sense of others’ behavior by internally simulating mental processes that, if enacted, would produce similar behaviors, such as intentional actions and emotional expressions.
Compared to other theories of mind, ST draws more heavily on biological evidence. It has been applied across fields such as cognitive science, neuroscience, developmental psychology, and clinical research.
In LLM research, ST forms a key foundation for AI empathy and other advanced social skills, and it received 113 citations from the surveyed LLM research papers. 
For instance, \citet{qian-etal-2023-harnessing} explored the role of ST in LLMs' empathetic responses, and \citet{shen-etal-2024-heart} conducted an empirical analysis with LLMs examining the relationship between empathy and narrative style in storytelling.
Their research has leveraged ST in empathy modeling, and \citet{nie2023moca} examined the influence of ST on how LLMs make correct moral judgment.
Furthermore, a sub-concept of ST called perspective-taking has been studied in LLMs. According to ST, it refers to the ability to understand a situation or concept from another person's point of view.
\citet{wilf-etal-2024-think} demonstrated that perspective-taking can effectively enhance LLMs' performance in mental state attribution, while \citet{xu-etal-2024-walking} incorporated it into prompting and highlighted its significance in reducing bias and toxicity in LLMs.
These findings indicate that ST may serve as a viable guide for developing more socially aware AI systems.

\textbf{Dual-Process Theory (DPT)} is a theory that explains how thought can arise in two distinct ways: System 1 and System 2. 
System 1 processes are implicit and unconscious, and may be influenced by persuasion or education;
System 2 processes are explicit and conscious, typically requiring more time to adapt to different situations.
These theories can be found in social, personality, cognitive, and clinical psychology, where the different modes of thinking are used to explain various phenomena. 
LLM researchers have primarily drawn on DPT to analyze and mitigate some social issues raised by these models, resulting in 103 citations among the surveyed LLM research papers.
For instance, \citet{sui-etal-2024-confabulation} referred to DPT when accounting for hallucinations in LLMs, and \citet{echterhoff-etal-2024-cognitive,koo-etal-2024-benchmarking} argued that both social and cognitive biases may stem from unconscious processes within LLMs, which parallel the System 1 processes described in DPT. 
Their work emphasizes the need for conscious audition to mitigate these problems.
Meanwhile, inspired by DPT, researchers have found it effective to influence patterns of thinking and behavior in LLMs through the use of personas.
\citet{sun-etal-2024-kiss} assigned different visual personas to MLLMs and observed corresponding behavioral changes.
\citet{hu-collier-2024-quantifying} quantified the impact of assigned personas on perspective simulation.
\citet{liu-etal-2024-evaluating-large} further evaluated the resulting social bias and steerability induced by different persona assignments.

\paragraph{Under-explored theories/frameworks}

We also list three other theories/frameworks that are closely related to the \textit{Social Cognition} cluster but have received relatively little attention in current LLM studies.

\textbf{Cognitive Dissonance Theory (CDT)} is a theory that focuses on the mental discomfort individuals experience when holding two or more conflicting cognitions, such as beliefs, attitudes, or behaviors. 
CDT posits that this dissonance motivates individuals to reduce the inconsistency, often by altering existing beliefs, justifying behaviors, or acquiring new information. 
CDT has been widely applied to understand processes like attitude change, decision-making, moral reasoning, and behavioral justification across various domains, such as consumer behavior, health psychology, and social dynamics.
In LLM research, CDT can be used as a theoretical framework for understanding user resistance to model suggestions, particularly when those suggestions conflict with users’ prior beliefs or intentions.
By modeling and anticipating dissonant reactions, LLMs can be designed to offer responses that reduce psychological discomfort, for example, by providing justifications, alternative framings, or gradual nudges toward behavior change.
Additionally, CDT can serve as an evaluation lens to assess how well LLM outputs align with users’ cognitive states and to measure whether interactions successfully reduce dissonance over time, thereby enhancing long-term trust and acceptance.

\textbf{Elaboration Likelihood Model (ELM)} is a theory of persuasion that explains how individuals process and respond to persuasive messages through two distinct routes, the central route and the peripheral route. 
The central route involves careful and thoughtful consideration of the message content, typically occurring when the individual is motivated and able to engage in deep cognitive processing.
In contrast, the peripheral route relies on superficial cues, such as the speaker’s credibility, attractiveness, or emotional appeal, when motivation or ability to process is low. 
ELM provides a comprehensive framework for understanding attitude change, highlighting that the durability and strength of persuasion depend on the route through which it is achieved. 
It has been widely applied in areas such as marketing, health communication, and political campaigning.
In LLM research, the ELM can serve as a framework for designing adaptive communication strategies based on user engagement levels.
By assessing users’ motivation and ability to process information, LLMs can tailor their responses to either follow the central route, providing detailed, logical arguments for highly engaged users; or the peripheral route, using concise, emotionally resonant cues for less engaged users.
This approach may ensure that model outputs are better aligned with users’ cognitive states.
Additionally, the ELM can be used as an evaluative lens to assess the quality and impact of LLM-generated persuasive content, guiding improvements in user experience and behavioral outcomes.

\textbf{Realistic Conflict Theory (RCT)} is a theory that explains intergroup conflict as arising from competition over limited resources. 
It posits that when groups perceive that they are in direct competition for resources such as jobs, power, or territory, hostility and prejudice are likely to emerge. 
RCT emphasizes the role of tangible, real-world conflicts of interest in generating negative intergroup attitudes and behaviors. 
RCT has been applied to understand a variety of social phenomena, including ethnic tensions, discrimination, and political polarization.
In LLM research, RCT may be used as a framework to analyze user interactions in competitive or zero-sum environments, such as online debates or resource allocation scenarios. 
By modeling how perceived intergroup threats influence communication patterns, developers can design LLMs that detect emerging conflicts and facilitate constructive dialogue. 
Additionally, RCT may inform the evaluation of LLM outputs in sensitive contexts by assessing whether responses exacerbate or mitigate perceived competition and group-based tensions.

\begin{table}
    \small
    \centering
    \begin{tabular}{ll}
    \toprule
    \textbf{Sub-Topic} & \textbf{Theory/Framework} \\
    \midrule
    \multirow{3}{*}{\begin{tabular}{l}
        Systems, Processes, and Brain Mechanisms of Memory
    \end{tabular}} & \cellcolor{myblue!22}Episodic vs. Semantic Memory \\
    & \cellcolor{myblue!22}Complementary Learning Systems \\
    & \cellcolor{myblue!15}Baddeley’s Model of Working Memory \\
    \midrule
    \multirow{2}{*}{\begin{tabular}{l}
        Science of Learning in Minds and Machines
    \end{tabular}} & \cellcolor{myblue!22}Complementary Learning Systems \\
    & \cellcolor{myblue!16}Bayesian Inference/The Bayesian Brain \\
    \midrule
    \multirow{3}{*}{\begin{tabular}{l}
        Developmental Neuroscience of Mind and Brain
    \end{tabular}} & \cellcolor{myblue!70}Executive Functions $\blacktriangleleft$ \\
    & \cellcolor{myblue!67}Theory of Mind $\blacktriangleleft$ \\
    & \cellcolor{myblue!14}Structure-Mapping Theory \\
    \midrule
    \multirow{4}{*}{\begin{tabular}{l}
        Reasoning, Analogy, and Theory of Mind
    \end{tabular}} & \cellcolor{myblue!67}Theory of Mind $\blacktriangleleft$ \\
    & \cellcolor{myblue!54}Theory-Theory $\blacktriangleleft$ \\
    & \cellcolor{myblue!34}Dual-Process Theory \\
    & \cellcolor{myblue!14}Structure-Mapping Theory \\
    \midrule
    \multirow{3}{*}{\begin{tabular}{l}
        Cognitive Science Theories of Mental Architecture
    \end{tabular}} & \cellcolor{myblue!67}Theory of Mind $\blacktriangleleft$ \\
    & \cellcolor{myblue!34}Dual-Process Theory \\
    & \cellcolor{myblue!14}Mental Model Theory of Reasoning \\
    \bottomrule
    \end{tabular}
    \caption{Subtopics and Corresponding Top Theories or Frameworks in the \textit{Neural Mechanisms} Cluster. \\ \textit{Note:} Cell opacity represents citation frequency; black triangles indicate the three most frequently cited theories/frameworks.}
    \label{tab:sub_4}
\end{table}

\begin{figure}
    \centering
    \includegraphics[width=\linewidth]{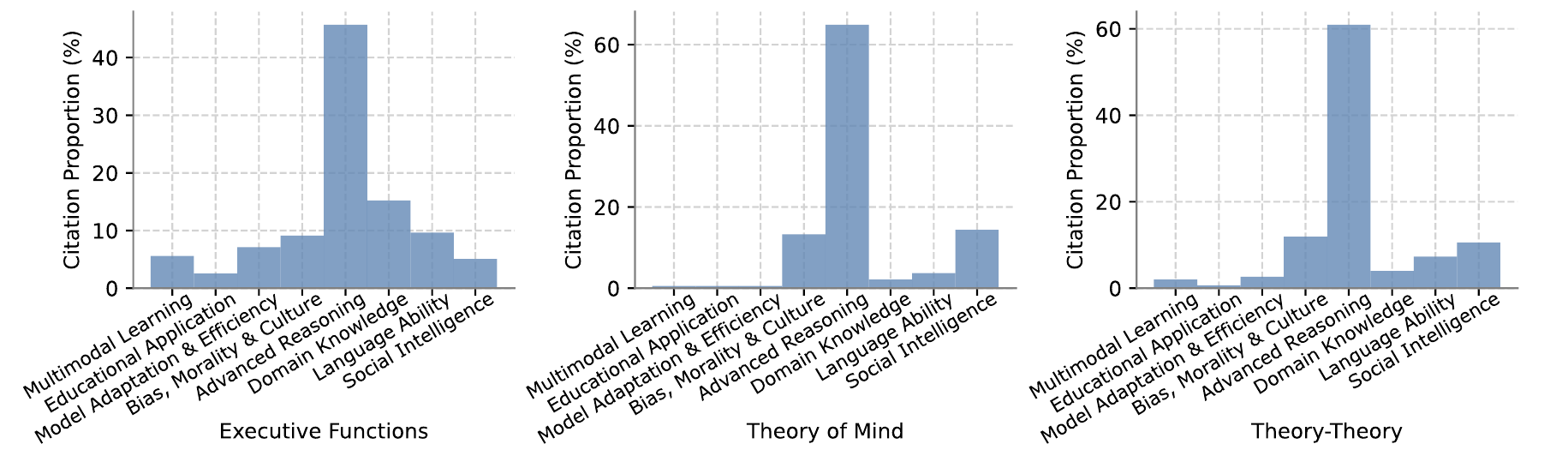}
    \caption{Citation distribution for the top three theories/frameworks in the \textit{Neural Mechanisms} cluster across eight LLM research topics.}
    \label{fig:theory_4}
    \vspace{-10pt}
\end{figure}

\subsubsection{Neural and cognitive mechanisms of learning and creativity}\label{sec:5.2.5}

\paragraph{Popular theories/frameworks}
In the \textit{Neural Mechanisms} cluster, the three most frequently referenced psychology theories/frameworks in the surveyed LLM research papers are \textit{Executive Functions}~\citep{Diamond2013Executive}, \textit{Theory of Mind}~\citep[ToM;][]{Premack1978-PREDTC-3,10.1093/biohorizons/hzq011}, and \textit{Theory-Theory}~\citep{Ratcliffe2006}.
Their citation distributions across the eight LLM research topics are shown in Fig.~\ref{fig:theory_4}.

\textbf{Executive Functions (EFs)} are a framework of cognitive processes that support goal-directed behavior, in which higher-order EFs require the coordinated use of multiple basic ones. 
All these functions develop gradually over the lifespan and can be improved at any point in a person’s life, though they may be adversely affected by various affective factors.
They play a fundamental role in people's actions and are deeply intertwined with domains such as mental health, social functioning, and academic achievement.
EFs have been widely studied in education, clinical psychology, neuroscience, workplace settings, and public health and policy, particularly in terms of how they can be improved and maintained, as well as how they contribute to various forms of goal-directed behavior.
EFs are the most frequently referenced theory/framework in the \textit{Neural Mechanisms} cluster.
LLM researchers have primarily drawn on EFs to enhance LLMs’ capabilities for corresponding behaviors, resulting in 197 citations among surveyed LLM research papers.
For example, higher-order EFs like problem-solving (e.g.,~\citealp{didolkar2024metacognitive,yao2023tree}) and planning (e.g.,~\citealp{hao2023reasoning,10.5555/3692070.3694316}) have become key areas in LLM reasoning.
Meanwhile, some foundational EFs, though less extensively studied, have also been explored as a means to support the development of higher-level capabilities in LLMs. 
For instance, both working memory’s essential role in LLMs’ reasoning abilities (e.g., \citet{wu2024minds,zhang-etal-2024-working}) and cognitive flexibility’s underpinning of their adaptive, context-sensitive behavior (e.g., \citealp{dong-etal-2023-steerlm,shao-etal-2023-character}) have been empirically demonstrated.
In addition to these main EFs, \citet{ren-xiong-2023-huaslim} leveraged attention control to inhibit irrational shortcut learning, thus enhancing models' generalization.

\textbf{Theory of Mind (ToM)}, the previously mentioned cognitive ability to attribute mental states, is supported by a dedicated network of brain regions and underlying EFs. 
Research on ToM in autism indicates that these specialized mechanisms can, in some cases, be selectively impaired while general cognitive function remains largely intact.
Neuroimaging studies further evidence the view, showing that the medial prefrontal cortex (mPFC), the posterior superior temporal sulcus (pSTS), the precuneus, and the amygdala are involved in ToM-related activities~\citep{tommechanisms}, such as social reasoning and decision-making.
Therefore, in addition to its role in social reasoning, ToM has been adopted in LLM research to guide the exploration and enhancement of underlying reasoning mechanisms.
In the psychology cluster of \textit{Neural Mechanisms}, the 188 citations of ToM are heavily concentrated in \textit{Advanced Reasoning} among the eight LLM research topics, exhibiting a distribution different from that shown in Fig.~\ref{fig:theory_3}.
A mainstream direction involves analyzing the mechanisms of LLM's social reasoning. For example, \citet{huang-etal-2024-notion} measured the complexity of different ToM tasks for LLMs, drawing on cognitive load theory.
\citet{10.5555/3692070.3694663} linearly decoded LLMs’ representations of their own and others’ belief states from neural activations, adopting a connectionist perspective, while \citet{jung-etal-2024-perceptions} evaluated the precursory inferences for ToM in LLMs to further develop their ToM abilities, adopting a symbolic perspective.
Another branch involves applying ToM to multi-agent interactions.
This includes LLM collaboration, where the ToMs of different agents are integrated through synergy into a unified system (e.g.,\citealp{li-etal-2023-theory,wang-etal-2024-unleashing}), 
and competition, where the ToMs of different agents are better differentiated and strengthened (e.g.,\citealp{10.5555/3692070.3692537,wu-etal-2024-deciphering,xu-etal-2024-magic}).

\textbf{Theory-Theory (TT)} is a theory concerning how humans develop an understanding of the outside world. 
While it shares with ToM the assumption that individuals possess a basic or naïve theory of psychology (i.e., folk psychology) to infer others’ mental states, TT is better understood as a broader framework for learning rather than a specific cognitive ability. 
It extends beyond reasoning about people and their viewpoints to include understanding mechanical devices and other non-agentive objects.
TT has been widely studied in developmental psychology, education, cognitive modeling, social psychology, and domain-specific reasoning.
From the perspective of TT, folk psychology, or the explanatory mental models in LLMs, is built through inductive reasoning. 
LLM researchers have primarily drawn on TT to enhance LLMs' capabilities in inductive reasoning and causal inference, resulting in 151 citations among the surveyed LLM research papers.
For example, \citet{wang2024hypothesis} proposed a pipeline for complex abstract hypothesis generation;
\citet{shani-etal-2023-towards} and \citet{suresh-etal-2023-conceptual} mirrored TT to interrogate LLMs' latent structure of conceptual representations, thereby achieving concept awareness;
and \citet{jiayang-etal-2023-storyanalogy} and \citet{wijesiriwardene-etal-2023-analogical} evaluated LLMs across text analogies at various levels, ranging from words and sentences to metaphors and stories.
These processes contribute to improved LLM performance in inductive reasoning tasks.
Furthermore, \citet{liu-etal-2023-magic} and \citet{nie2023moca} drew on TT when investigating LLMs’ ability to derive cause-effect relationships.

\paragraph{Under-explored theories/frameworks}

We also list three other theories/frameworks that are closely related to the \textit{Neural Mechanisms} cluster but have received relatively little attention in current LLM studies.

\textbf{Levels of Processing Model (LPM)} is a framework that posits memory retention is influenced by the depth at which information is processed. 
Rather than focusing on separate memory stores, the model emphasizes the continuum of processing levels, ranging from shallow (e.g., perceptual or structural features) to deep (e.g., semantic meaning and personal relevance). 
Deeper levels of processing lead to more durable and accessible memory traces. 
LPM has been influential in understanding encoding mechanisms and has implications for educational practices, memory enhancement strategies, and interventions for memory-related disorders.
In LLM research, LPM holds great potential for enhancing the general learning process. 
Whether a similar mapping between levels of processing and memory duration exists in LLMs is worth investigating.
If so, the three levels of processing (i.e., structural/visual, phonemic, and semantic) correspond to three modalities, which may inspire new learning algorithms for MLLMs. 
For example, LPM may serve as a guide for data administration at various training stages to mitigate forgetting or imbalance across different modalities.
Moreover, if we further abstract the mapping in LPM, a conceptual parallel may emerge between the depth of human processing and the layers in deep neural models. 
By leveraging LPM-like mappings as heuristics in model adaptation, more PEFT methods become possible from an information processing perspective.
Meanwhile, from a memory duration perspective, the personalization of LLMs can be effectively managed to balance steerability and stability.

\textbf{Piaget’s Stage Theory of Cognitive Development (PSTCD)} is a theory that outlines how children's thinking evolves through a series of qualitatively distinct stages, each characterized by different cognitive abilities. 
PSTCD posits that children actively construct knowledge as they interact with their environment, progressing through four stages: sensorimotor, preoperational, concrete operational, and formal operational. 
Each stage represents a shift in how children understand and engage with the world, highlighting the importance of maturation and experience in cognitive growth. 
PSTCD has significantly influenced educational practices and our understanding of child development.
In LLM research, PSTCD may serve as a valuable conceptual framework for modeling human-like learning trajectories. 
By incorporating the stage-based characteristics of cognitive development, researchers may design training curricula that progress from simple, concrete tasks to more abstract, logical reasoning, mirroring the natural evolution of human cognition. 
Furthermore, aligning interaction strategies with different cognitive stages allows LLMs to generate age-appropriate and educationally tailored responses, making them more effective for personalized learning environments. 
PSTCD also offers a structured lens for evaluating a model's cognitive maturity, guiding the design of benchmarks that reflect developmental reasoning skills.

\textbf{Hebbian Theory (HT)} is a theory that emphasizes the role of synaptic plasticity in learning and memory. 
HT posits that the repeated and persistent activation of one neuron by another strengthens the connection between them, thereby shaping neural networks.
HT has been influential in understanding brain development, learning processes.
In LLM research, HT may offer valuable inspiration for understanding and designing learning mechanisms. 
HT's core idea, neurons that fire together, wire together, has influenced the development of neural networks by introducing local learning rules and concepts of synaptic plasticity. 
These ideas may support the exploration of more biologically plausible and interpretable models. 
Furthermore, HT emphasizes associative memory, may offer insights into how LLMs might enhance their capacity for long-term memory and contextual association. 
This may have implications for future architectures, such as neuromorphic computing and memory-augmented models, where energy-efficient and adaptive learning processes are increasingly important.

\begin{table}
    \small
    \centering
    \begin{tabular}{ll}
    \toprule
    \textbf{Sub-Topic} & \textbf{Theory/Framework} \\
    \midrule
    \multirow{3}{*}{\begin{tabular}{l}
        Survey Design, Experimentation and Science \\
        Communication
    \end{tabular}} & \cellcolor{mypurple!70}Dual-Process Theory $\blacktriangleleft$ \\
    & \cellcolor{mypurple!15}Cognitive Aspects of Survey Methodology \\
    & \cellcolor{mypurple!2}Cultural Consensus Theory \\
    \midrule
    \multirow{3}{*}{\begin{tabular}{l}
        Measurement and Application of Psychometrics
    \end{tabular}} & \cellcolor{mypurple!43}Classical Test Theory $\blacktriangleleft$ \\
    & \cellcolor{mypurple!8}Item Response Theory \\
    & \cellcolor{mypurple!4}Multitrait-Multimethod Matrix \\
    \midrule
    \multirow{3}{*}{\begin{tabular}{l}
        Bias and Irrationality in Human Judgment
    \end{tabular}} & \cellcolor{mypurple!62}Heuristics and Biases Program $\blacktriangleleft$ \\
    & \cellcolor{mypurple!27}Rational Choice Theory/Game Theory \\
    & \cellcolor{mypurple!1}Fuzzy-Trace Theory \\
    \midrule
    \multirow{3}{*}{\begin{tabular}{l}
        Cognitive Models of Human Reasoning and \\
        Decision-Making
    \end{tabular}} & \cellcolor{mypurple!62}Heuristics and Biases Program $\blacktriangleleft$ \\
    & \cellcolor{mypurple!14}Causal Models/Causal Bayes Nets \\
    & \cellcolor{mypurple!4}Evolutionary Psychology Approach to Reasoning \\
    \bottomrule
    \end{tabular}
    \caption{Subtopics and Corresponding Top Theories or Frameworks in the \textit{Psychometrics \& JDM} Cluster. \\ \textit{Note:} Cell opacity represents citation frequency; black triangles indicate the three most frequently cited theories/frameworks.}
    \label{tab:sub_5}
\end{table}

\begin{figure}
    \centering
    \includegraphics[width=\linewidth]{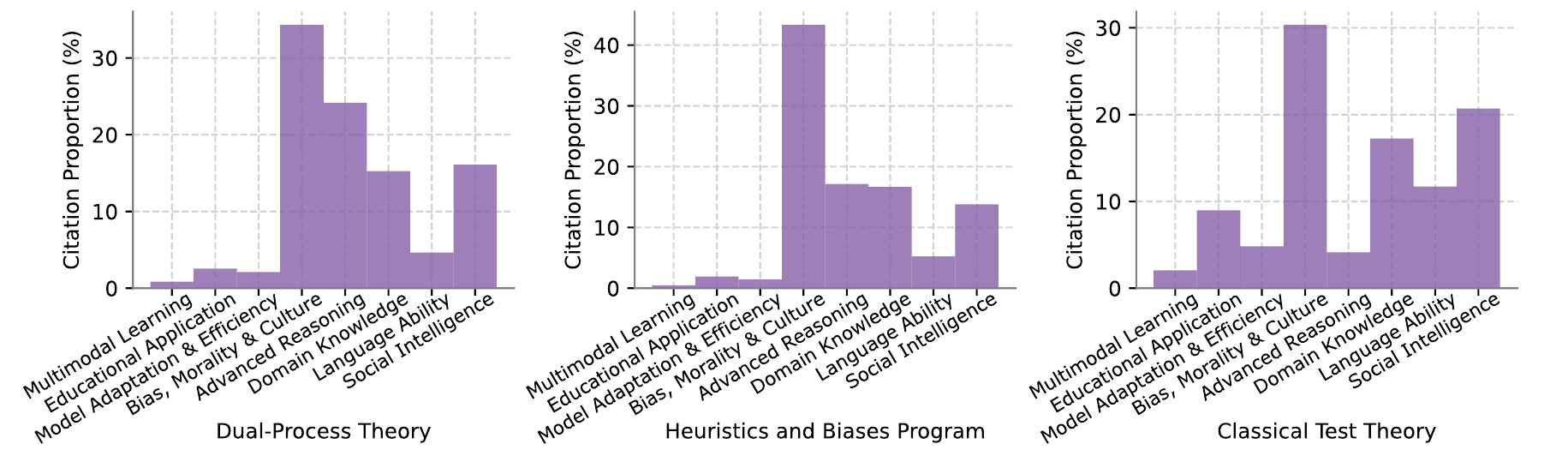}
    \caption{Citation distribution for the top three theories/frameworks in the \textit{Psychometrics \& JDM} cluster across eight LLM research topics.}
    \label{fig:theory_5}
\end{figure}

\subsubsection{Psychometrics, and judgment and decision-making}\label{sec:5.2.6}

\paragraph{Popular theories/frameworks}
In the \textit{Psychometrics \& JDM} cluster, the three most frequently referenced psychology theories/frameworks in the surveyed LLM research papers are \textit{Dual-Process Theory}~\citep{gawronski2013dpt}, \textit{Heuristics-and-Biases Program}~\citep{doi:10.1126/science.185.4157.1124}, and \textit{Classical Test Theory}~\citep[CTT;][]{lord1968statistical,NOVICK19661}.
Their citation distributions across the eight LLM research topics are shown in Fig.~\ref{fig:theory_5}.

\textbf{Dual-Process Theory (DPT)}, previously mentioned as the theory that partitions human cognition into two distinct types of processes, has significantly influenced studies of executive control, reward-based learning, and judgment and decision-making.
According to some researchers, System 1 and System 2 do not operate as parallel systems~\citep{thinkingstraight}. 
Typically, System 1 generates intuitive responses, which are then monitored and evaluated by System 2. However, System 2 does not always override System 1, especially under time pressure, cognitive load, or distraction.
This has inspired LLM researchers to propose that not only social interactions but also a broader range of LLMs' behaviors can be moderated and steered through DPT and related frameworks.
In the \textit{Psychometrics \& JDM} cluster, DPT is the most frequently cited psychology theory/framework, with 236 citations across the surveyed LLM research papers. 
For example, \citet{yao2023tree} incorporated planning processes into general problem-solving, \citet{dziri2023faith} used computation graphs for compositional tasks, and \citet{xu-etal-2024-faithful} applied chain-of-thought reasoning in logical reasoning.
Their research improves LLMs' decision-making by introducing a conscious, explicit guide to assist the unconscious, greedy processes.
This comparison of LLMs' original cognitive processes to \textit{System 1} in DPT is supported by evidence that LLMs can be easily persuaded (e.g.,~\citealp{xie2024adaptive,xu-etal-2024-earth}).
\citet{goldstein-etal-2023-decoding} provides further evidence that LLMs’ limited verification capability corresponds to \textit{System 2}.
Given this similarity, the trade-off between the two systems has invited discussions about consistency and uncertainty in LLMs (e.g.,~\citealp{jang-lukasiewicz-2023-consistency,yona-etal-2024-large}), particularly when dealing with knowledge conflicts.

\textbf{Heuristics and Biases Program (HBP)} is a research framework that investigates how people rely on heuristics to make decisions under uncertainty, and how these heuristics can lead to systematic errors or cognitive biases.
Heuristics refers to the process by which humans use mental shortcuts to quickly arrive at judgments, decisions, and even solutions to complex problems.
In the early 1970s, it became closely associated with cognitive biases through a series of experiments, demonstrating that people’s intuitive judgments often deviated from normative rules.
HBP informs research and practice in behavioral economics, clinical psychology, law, public policy, education, and organizational management.
LLM researchers have mainly drawn on HBP to reveal how heuristics shape decision-making and how to mitigate related biases.
HBP received 210 citations from the surveyed LLM research papers, most of which fall within the \textit{Bias, Morality \& Culture} cluster.
For instance, \citet{echterhoff-etal-2024-cognitive} identified cognitive biases in LLMs under high-stakes scenarios and proposed a strategy for the models to mitigate their own human-like biases; \citet{jiang-etal-2024-peek} assessed whether LLMs possess genuine reasoning abilities or primarily depend on token bias.
On the other hand, heuristics have been shown to be valuable for boosting the efficiency of LLMs in search (e.g.,~\citealp{gupta-li-2024-training,yao2023tree}) and reasoning (e.g.,~\citealp{bertolazzi-etal-2024-systematic,pan-etal-2024-dynathink}). 
These heuristic approaches either mimic human heuristics or rely on computational strategies to balance performance and cost, while \citet{zhou-etal-2024-llms} also validated that LLMs can automatically acquire task-specific heuristics from in-context demonstrations.

\textbf{Classical Test Theory (CTT)} is a psychometric theory concerned with predicting outcomes of psychological tests, such as item difficulty and examinee ability. 
It is based on the idea that a person’s observed score on a test is the sum of a true score (the error-free score) and an error score. 
The aim of CTT is to understand and improve the reliability of psychological assessments, that is, to ensure test scores are precise, reproducible, and consistent across different testing conditions.
LLM researchers have extensively adopted CTT in the design of evaluation methods, resulting in 145 citations among the surveyed papers.
Many existing LLM evaluation methods \textit{de facto} follow the logic of CTT implicitly, testing models with a range of items and reporting average scores (e.g.,~\citealp{manakul-etal-2023-selfcheckgpt,10.5555/3666122.3668142}). 
This is due to the straightforward assumptions of CTT, including the decomposition into true scores and errors, the linear and additive nature of the model, and the homogeneity of test items.
Nonetheless, only a little research (e.g.,~\citealp{cao-etal-2024-structeval,forde-etal-2024-evaluating,li-etal-2024-split}) formalizes error variance or true score modeling as CTT would.
Moreover, some research has indicated that CTT would be less reliable for evaluating LLMs given its simple assumptions. 
For example, \citet{xiao-etal-2023-evaluating-evaluation} observed noise in true score and error modeling, and proposed a testing framework to measure both the reliability and validity of NLG metrics.
\citet{kobayashi-etal-2024-revisiting} proposed a benchmark that combines test items of varying evaluation granularity, aiming to mitigate inconsistencies across different tests.
The context-sensitivity of LLMs and the nuanced nature of test items motivate the development of more context-aware, item-level, and dynamic measurements.

\paragraph{Under-explored theories/frameworks}

We also list three other theories/frameworks that are closely related to the \textit{Psychometrics \& JDM} cluster but have received relatively little attention in current LLM studies.

\textbf{Lincoln and Guba's Evaluative Criteria (LGEC)} is a framework for assessing the trustworthiness of qualitative research. 
LEGC emphasizes the importance of rigor through four key dimensions: credibility, transferability, dependability, and confirmability. 
These criteria serve as qualitative counterparts to the concepts of internal validity, external validity, reliability, and objectivity in quantitative research. 
The framework aims to ensure that qualitative findings are both believable and applicable, offering researchers a systematic approach to evaluating and enhancing the quality of their work.
In LLM research, LGEC may guide the evaluation of model-generated qualitative outputs, such as LLM-as-judge for narrative responses, user reflections, or topic modeling.
By applying these criteria, researchers can assess whether the responses produced by LLMs are contextually relevant (transferable), logically consistent (dependable), and grounded in source data or reasoning (confirmable), thereby enhancing the trustworthiness and practical value of AI-driven qualitative analysis.

\textbf{Prospect Theory (PT)} is a theory that describes how individuals make decisions under conditions of risk and uncertainty, highlighting the psychological biases that diverge from rational choice. 
PT emphasizes that people evaluate potential losses and gains relative to a reference point, and that losses typically loom larger than equivalent gains (known as loss aversion). 
PT has been instrumental in explaining real-world decision-making patterns in areas such as finance, consumer behavior, and public policy.
In LLM research, PT offers valuable insights for modeling and simulating human decision-making under uncertainty. 
By incorporating key principles such as loss aversion and reference dependence, LLMs can better reflect the psychological biases that influence human judgment. 
This is particularly useful in areas like dialogue generation, recommendation systems, and behavior prediction, where understanding users' risk preferences enhances personalization and realism. 
PT may also inform the design of reward functions in reinforcement learning from human feedback, enabling LLMs to align more closely with real-world human values and sensitivities.
Furthermore, PT-guided framing strategies may improve the persuasive impact of generated content in domains like marketing, public policy, and healthcare communication.

\textbf{Framing Theory (FT)} is a theory that explores how information presentation influences individuals’ perception and interpretation of events, issues, and messages. 
FT emphasizes the role of context, language, and emphasis in shaping meaning, guiding attention, and influencing emotional and behavioral responses. 
By highlighting certain aspects of a message while downplaying others, framing can significantly affect public opinion, decision-making, and social discourse.
FT has been widely applied in media studies, political communication, public health campaigns, and social movement research.
In LLM research, FT may offer valuable insights into how the presentation of prompts, context, and language influences model outputs. 
By highlighting the importance of emphasis, wording, and contextual cues, FT helps explain why different phrasings of the same question can lead to significantly varied responses from a model.
This has implications for prompt engineering, bias detection, and user experience design. 
FT also aids in analyzing the latent frames within training data, which may introduce subtle biases into model behavior. 
Furthermore, FT may help researchers to explore public reactions under various narrative styles and assess emotional tone. 
Such as research on persuasive strategies in human-AI interaction can draw on this theory to examine how language framing influences user compliance and perception.
\subsection{How is psychology research operationalized and interpreted in the context of LLM research?}\label{sec:5.3}

In §\ref{sec:5.2}, we provided an overview and analysis of the psychology theories/frameworks cited in LLM research. 
Building on that foundation, this section further explores how LLM research concretely operationalizes and interprets the psychology literature, theories, and frameworks it references. 
Unlike the more macro-level overview in §\ref{sec:5.2}, the focus here is on the specific ways LLM research applies these psychological insights in practice, including potential misapplications or oversights.

Given the varying theoretical depth and scope of different psychology research, it is inevitable that LLM research exhibits diverse approaches to its application. 
Therefore, we adopt a case study approach to closely examine how the particular theory/framework is cited and used. 
Considering that ToM is one of the most commonly referenced psychology concepts in LLM research, we select ToM as a central case for in-depth analysis.

Although LLM research displays considerable variation in how it references different psychology theories, there are common patterns of misapplication in operationalization and theoretical understanding.
Through the analysis of the ToM case, we aim to reveal these shared issues and offer insights for more rigorous and accurate incorporation of psychology research into LLM studies.

\subsubsection{Case study: Theory of Mind}\label{sec:5.3.1}

As introduced in §\ref{sec:5.2}, Theory of Mind (ToM) refers to the capacity of individuals to attribute mental states such as beliefs, intentions, knowledge, and emotions to others, recognizing that these states may differ from their own.
This concept has become an important focus in LLM research, as it reflects the models’ potential to comprehend and reason about complex human mental states, which is considered essential for the development of more advanced and human-like artificial intelligence systems.
As discussed in §\ref{sec:5.2}, ToM is among the three most influential theories/frameworks cited in LLM research, drawing from the \textit{Social Cognition} and \textit{Neural Mechanisms} clusters within psychology research.
We conducted a comprehensive analysis of LLM papers that reference ToM-related works from these two clusters.

\begin{figure}
    \centering
    \includegraphics[width=0.8\linewidth]{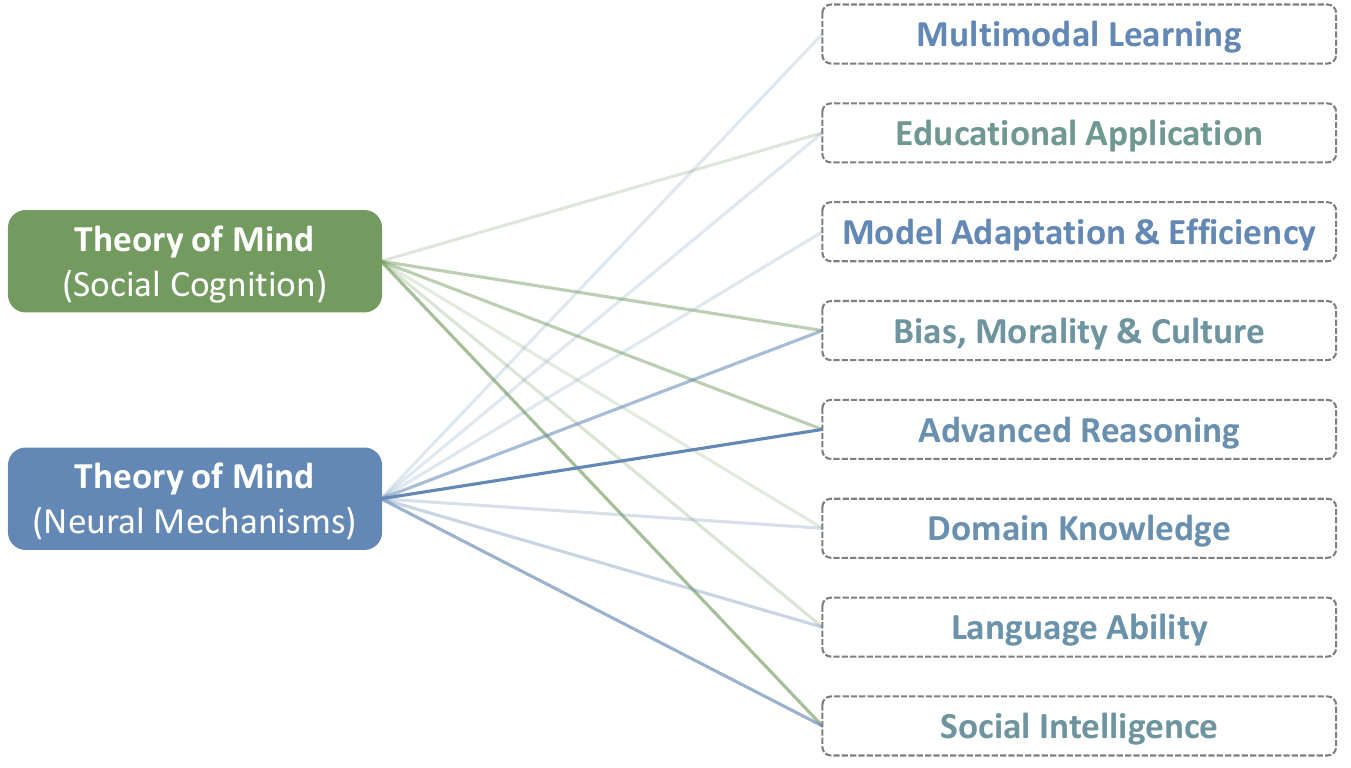}
    \caption{Bipartite Network of Citations Linking LLM Research Papers to Psychology Papers on Theory of Mind. \\ \textit{Note:} Line opacity reflects citation frequency, and the color of LLM research topics indicates the proportion of citations to psychology papers in the two clusters.}
    \label{fig:case_tom}
\end{figure}

ToM-related psychology papers within the \textit{Social Cognition} and \textit{Neural Mechanisms} clusters show clear differences in research orientation and methodological approaches. 
Research in the \textit{Social Cognition} cluster primarily focuses on the role of ToM in social interactions, emphasizing its relationships with abilities such as emotion understanding, social reasoning, empathy, and moral judgment. 
These studies often employ behavioral experiments, questionnaires, or situational tasks, highlighting the influence of developmental processes and social environmental factors on ToM.
In contrast, research in the \textit{Neural Mechanisms} cluster is more concerned with the biological foundation of ToM, exploring activation patterns in relevant brain regions (such as the prefrontal cortex and the temporoparietal junction) during ToM tasks. 
These studies commonly use techniques like functional magnetic resonance imaging (fMRI) and electroencephalography (EEG), following a neuroscience-oriented paradigm and focusing more on revealing the neural structure and functional mechanisms of ToM.
Therefore, the way LLM research references and uses ToM-related papers from different clusters also tends to differ.

When LLM research references studies from the \textit{Social Cognition} cluster, it primarily draws on established experimental tasks and research paradigms from that field as tools to evaluate the model's ``ToM-like'' abilities.
For example, researchers often borrow tasks like the False Belief Task (e.g., the classic Sally-Anne test) to assess whether a model can distinguish between a character's perspective and reality. 
Other common tests like the Smarties test (first-order) and the Ice Cream Van test (second-order) are also used to evaluate whether the model can maintain mental state modeling across multiple dialogue turns.
Additionally, some LLM studies adapt situational attribution tasks, narrative comprehension tasks, and social reasoning tasks to examine whether a model can grasp implicit intentions, emotional shifts, or social norms of characters. 
These tasks offer a structured and comparable framework for assessing LLMs.

When LLM research references studies from the \textit{Neural Mechanisms} cluster, the primary focus is on how ToM is supported at the neural level, as well as the mechanisms and biological foundations underlying how some individuals process others' mental states.
This includes examining the functional differentiation of brain regions such as the prefrontal cortex and temporoparietal junction in tasks like attribution, behavior prediction, and emotion understanding.
Inspired by these studies, LLM researchers design new architectures for LLMs, develop multi-agent systems/frameworks, and seek to explain model behaviors.
For example, some LLM researchers introduce processes analogous to ToM by incorporating concept-level representations during training, which they want to allow models to develop an understanding of concepts prior to engaging in tasks like language generation or comprehension.
Another example is LLM researchers inspired by the structure of social cognitive networks in neuroscience to structure the multi-agent systems, where each agent simulates the mental state of a specific role.

\paragraph{Common misapplication} 
In addition to the above discussion on how ToM is used in LLM research, we have identified four common types of misuse when citing ToM-related papers. 
Although these misuses specifically occur in the context of referencing ToM works, they also reflect broader issues in how current LLM research draws on findings from psychology research.

\begin{itemize}[topsep=0pt,left=8pt,itemsep=0pt]
\item \textbf{Conceptual overgeneralization and misclassification} 
One of the most common misuses in LLM research when citing ToM-related papers is the overgeneralization and misclassification of the ToM concept.
Researchers often reference psychology research on ToM without sufficiently attending to the original studies' specific research designs, populations, and experimental conclusions. 
As a result, ToM is frequently treated as a catch-all label, with key distinctions between task types and cognitive processes overlooked.
Ideally, when citing a paper, researchers should clearly convey the study's core findings and scope of applicability, then thoughtfully relate these to the observed behaviors or capacities of LLMs. 
However, many LLM studies adopt ToM-related terminology, such as belief reasoning, perspective-taking, or false belief tasks, without a deep understanding of the cited studies.
A common example of such conceptual overgeneralization is the failure to distinguish between different levels of mental-state reasoning tasks, such as first-order and second-order ToM, referring to them broadly as ``ToM tasks'' without clarifying their differing cognitive demands and underlying psychological mechanisms. 
For instance, some studies treat the Sally-Anne task (a first-order false belief task) and the Ice Cream Van task (a second-order false belief task) as equivalent tests of ToM when evaluating whether LLMs possess ToM abilities. 
Yet in psychology, these tasks are clearly differentiated: the former requires understanding that someone can hold a false belief about the world, while the latter involves reasoning about someone's belief about another person's belief (i.e., a recursive representation of mental states).
LLM researchers need to reflect more deeply on why a particular paper is being cited over others, and what specific conclusions or experimental designs from that study meaningfully contribute to LLM research.

In addition, there is a tendency in some LLM research to mistakenly categorize psychology studies that were not originally intended to explore ToM as supporting evidence for ToM. 
For example, studies focusing on emotion recognition, social attention, or attribution mechanisms are sometimes included in discussions of ToM. 
Although these studies are indeed related to social cognition, they do not strictly fall within the core components of ToM, especially when they do not involve recursive belief construction or the understanding of mental states. 
For example, emotion recognition and belief attribution are distinct psychological processes. 
The former relies more on perception, while the latter involves recursive modeling of mental states. 
This practice may stem from a vague understanding of what constitutes ToM, or from a selective interpretation of psychology research when constructing arguments, thereby weakening the conceptual rigor of ToM in interdisciplinary research.

\item \textbf{Partial or incomplete citation} 
Another common misuse is that when citing psychology research on ToM, researchers often only select a few well-known studies while overlooking other equally important but less ``representative'' work in the field. 
We understand that citing well-known papers helps strengthen the credibility and authority of a paper, especially in literature reviews or theoretical frameworks, where referencing widely recognized classic studies can build a solid academic foundation.
However, this practice can also lead to a narrow research perspective and neglect the diversity of findings within the field.
In fact, some less ``representative'' studies, although less well-known, still offer valuable contributions that complement, challenge, or deepen mainstream views through their methods, samples, or conclusions. 
These can be especially useful as references for certain types of LLM research on ToM.
For example, most LLM studies referencing psychology research on ToM often cite~\citet{Premack1978-PREDTC-3} work on whether chimpanzees possess ToM, or~\citet{BARONCOHEN198537} study on ToM in children with autism.
Although these studies are undoubtedly milestones in ToM research history, they may not always be the most suitable for exploring ToM in the context of LLMs. 
Specifically, some LLM-related studies want to examine how reasoning abilities evolve across different training stages or attempt to break down the model's ``ToM-like'' capabilities into multiple levels for analysis.
In this context, the study by~\citet{wellman2004scaling} on the developmental progression of ToM may offer more relevant insights. 
Their research designed a set of hierarchically structured tasks to demonstrate that children's ToM abilities develop gradually through several stages. 
This phased, progressive perspective aligns more closely with how LLMs' capabilities are described and offers a more structured theoretical framework for refining the analysis of ``ToM-like'' abilities.

Moreover, this emphasis on well-known research has led LLM research to primarily draw on a limited number of widely cited experimental findings, while overlooking many studies that also hold significant theoretical value and experimental insight. 
For example, \citet{apperly2006belief} found that even adults do not automatically engage ToM abilities in certain contexts; instead, they rely on cognitive control resources to perform reasoning tasks about mental states.
This finding can be important for understanding the conditions under which LLMs might exhibit an active use of ToM-like abilities. 
Similarly, \citet{onishi200515}, using the violation of expectation paradigm with infants, demonstrated that sensitivity to others' belief states may emerge at an earlier developmental stage than previously thought.
This provides empirical support for the concept of ``early ToM'' and offers valuable clues for exploring whether LLMs might develop some form of ToM processing during pretraining. 
Although these less frequently cited studies are not as widely known as classic experiments by \citet{BARONCOHEN198537} or \citet{Premack1978-PREDTC-3}, they offer unique value to current LLM research in terms of methodology, theoretical perspective, and task design. 
Continued neglect of such studies in LLM research risks missing important opportunities to deepen our understanding of ToM-like capabilities in LLM research.

\item \textbf{Misinterpretation or misrepresentation of findings} 
Misinterpretation of cited psychology research is also a common issue in LLM research. 
This may have led some LLM studies to cite inappropriate papers to support certain arguments. 
This problem may stem from a lack of disciplinary sensitivity among LLM researchers when dealing with interdisciplinary literature, which may result in the use of studies that appear relevant on the surface but fail to provide adequate support.
For example, some LLM research cites psychology studies on the biological mechanisms of ToM to support discussions about ToM performance at the social level. 
This conflates theoretical frameworks and research goals across different levels. 
The activation of a specific brain region may indicate how a certain function is recruited, but it cannot directly show the individual's behavioral strategies or interaction styles in specific social contexts.
Some researchers may use the superficial criterion of ``this article studied ToM'' to include it as supporting material, thereby masking logical leaps in their argumentation.

In addition, some LLM research selectively emphasizes positive findings while overlooking the limitations of the psychology research they cite.
For example, researchers may highlight only the conclusive statements from cited research, while ignoring important caveats noted by the original authors, such as experimental boundaries, sample limitations, or theoretical controversies. 
In some cases, even papers that have been widely questioned (have controversial opinions) are cited without any clarification.
This kind of selective referencing can lead to a biased evidentiary base, thereby weakening the rigor of arguments.
For example, some papers refer to the two-systems account of ToM and observational studies on children’s gaze behaviors to support the idea that LLMs may possess implicit ToM abilities, yet they fail to adequately address the ongoing debates and limitations within these studies. 
Notably, there is no scholarly consensus on whether children’s gaze-shifting behavior truly indicates an implicit ToM.
Some researchers suggest that such behaviors may stem from low-level attentional preferences rather than genuine mental state attribution. 
We recommend that LLM researchers adopt a critical mindset when referencing psychology research, presenting the theoretical context, methodological constraints, and academic debates of the cited studies in a balanced way to ensure accuracy and scientific rigor in their arguments.

\item \textbf{Secondary citation errors} 
Another characteristic of how LLM research cites ToM papers is that researchers tend to rely more on secondary literature within the AI community when constructing ToM-related arguments, rather than directly consulting or citing primary literature from the field of psychology.
In other words, many theoretical claims or experimental justifications concerning ToM are based on interpretations by other LLM/NLP researchers rather than on direct engagement with original psychology research.
This phenomenon is not uncommon in scientific communication and within knowledge communities, and it is easy to understand why: researchers are more inclined to cite well-recognized work within their own field to enhance the acceptability and persuasiveness of their arguments during submission, peer review, or academic evaluation.
This ``in-group citation preference'' tends to create a closed citation loop. 
Early studies that introduced ToM into the LLM context established a framework of terminology and tasks, and subsequent research continues to build upon this framework, gradually forming an internally-reinforced citation network.

However, the cumulative bias inherent in secondary citations can further amplify the previously mentioned issues, such as conceptual overgeneralization, task confusion, narrow literature selection, and misinterpretation.
If one psychology research is misunderstood during its initial citation, subsequent literature that continues to cite this interpretation without verification can lead to a ``consensus of misreading.''
More critically, the nuanced descriptions, methodological complexities, and theoretical debates surrounding ToM in original psychology research are often compressed, oversimplified, or even omitted in secondary citations.
For example, some papers may not directly engage with the original studies by~\citet{BARONCOHEN198537} on autism and ToM, but instead rely on brief summaries from other LLM research, which can risk overlooking essential elements such as experimental controls and sample differences.
We recommend that LLM researchers trace psychology theories or experiments back to their original sources, rather than relying solely on interpretative summaries from the NLP/LLM community. 
Only by returning to the original citations can one clarify the theoretical context, research intent, and methodological limitations, thereby ensuring accuracy in understanding and rigor in application.
\end{itemize}

\section{Discussion}\label{sec:6}

\subsection{Summary of key findings}\label{sec:6.1}
Here, we summarize the answers to the three research questions posed at the end of §\ref{sec:1}.

\textbf{RQ1}: How is psychology research integrated into LLM research?

\textbf{A1}: Since 2023, an increasing number of LLM research papers have cited psychology papers, indicating a growing interest among researchers in insights from psychology. 
So far, psychology has been broadly integrated into LLM research, with \textit{Neural Mechanisms} and \textit{Psychometrics \& JDM} being the most prominent psychology topics.
LLM research topics demonstrate distinct referencing preferences for different areas of psychology. 
For example, \textit{Educational Application} and \textit{Advanced Reasoning} clearly favor psychology papers from the \textit{Education} and \textit{Neural Mechanisms} clusters, respectively, whereas \textit{Model Adaptation \& Efficiency} and \textit{Social Intelligence} draw on a much broader range of psychology topics.
These citation patterns result from the interplay between the nature of research topics in psychology and LLM research, as well as the characteristics of LLMs, such as being fast-updating, data-intensive, and black-box. More detailed discussions can be found in §\ref{sec:5.1}.

\textbf{RQ2}: Which psychology theories/frameworks are most commonly used, and which remain
underexplored in LLM research?

\textbf{A2}: The top 10 psychology theories and frameworks most frequently cited by the surveyed LLM research papers are Dual-Process Theories (434 citations), Theory of Mind (309 citations), Heuristics and Biases Program (210 citations), Executive Functions (197 citations), Connectionism vs. Symbolism (190 citations), Theory-Theory (151 citations), Classical Test Theory (145 citations), Usage-based Models of Language (128 citations), Mental Simulation Theory (113 citations), and Schema Theory (109 citations).
The application of these theories and frameworks reveals that the LLM research paradigm is becoming increasingly pluralistic under the influence of psychology. 
This paradigm begins as performance-driven and model-centric. As more psychology theories and frameworks are incorporated to guide experiments on LLMs, it gradually adopts a theory-driven and data-centric, i.e., empirical, approach.
However, despite the many theories and frameworks within each psychology topic, as elaborated on in §\ref{sec:5.2}, only some have been engaged by LLM research, leaving ample room for further exploration.

\textbf{RQ3}: How is psychology research operationalized and interpreted in the context of LLM research?

\textbf{A3}: A psychology theory/framework has multiple facets that can be studied from different perspectives in psychology research, leading to a variety of applications in LLM research that cite them. 
A case study on Theory of Mind is presented in §\ref{sec:5.3} to exemplify this diversity.
Across the different methods of operationalization and interpretation, there are four types of common misapplications: 
1) \textit{Conceptual Overgeneralization and Misclassification}, where LLM researchers cite related psychology research clarifying its primary design, target population, or key conclusions; 
2) \textit{Partial or Incomplete Citation}, where LLM researchers rely on a few popular papers about their intended psychology theories and frameworks, overlooking other, potentially more relevant, but less well-known research;
3) \textit{Misinterpretation or Misrepresentation of Findings}, where LLM researchers cite inappropriate psychology papers to support their arguments or overly emphasize partial findings from the cited papers, despite only topical relevance; and
4) \textit{Secondary Citation Errors}, where LLM researchers prefer citing influential LLM research papers that engage with the intended psychology theories and frameworks over the original psychology research itself.
All these misuses, while they may lead to surprising findings, could compromise the validity and accuracy of insights drawn from psychology.

\subsection{Theoretical and methodological reflections}\label{sec:6.2}

Although the intersection of AI research and psychology is advancing rapidly, productive interdisciplinary integration continues to face theoretical and methodological challenges.

First, on the theoretical level, current AI research often tends to instrumentalize psychology theories, simplifying complex ideas into quantifiable conceptual labels.
Although this simplification lowers the threshold for applying theories, it can also obscure the deeper structures of the original frameworks.
For example, the use of ToM in AI research often treats mental states like beliefs and desires as static data points.
In contrast, within psychology research, ToM is understood as a dynamic, context-dependent ability, typically characterized by uncertainty, ambiguity, and developmental variability.
Behind this simplification lies a fundamental difference between the goals of psychology theories and the goals of AI researchers when applying these theories. 
Psychology aims to explain the internal mechanisms of human behavior and mental activity, emphasizing the complexity of processes and the importance of social and cultural contexts.
In contrast, AI researchers tend to focus more on functional reconstruction and engineering implementation, often transforming psychology concepts into operational and computable model parameters.
The risk of such oversimplification in AI can lead to misleading conclusions about the capabilities of AI systems, such as overestimating their understanding of human behavior or wrongly attributing human-like intentionality to models that merely simulate behavioral patterns. 
By failing to account for the fluid, interpretive nature of social reasoning, AI models risk reinforcing shallow imitations of human cognition, which may perform well in isolated tasks but lack the flexibility and depth needed for genuinely adaptive or ethical interaction.
To avoid conceptual drift, AI researchers referencing psychology theories should pay close attention to their original context, like core assumptions and theoretical boundaries, rather than merely adopting surface-level terminology.

Additionally, psychology theories are often used as tools to explain model behavior, but such post hoc explanations often lack predictive power and systematic structure. 
One example is to use attention in cognitive psychology to explain the behavior of neural models like Transformers.
This explanation is derived after observing the model’s output and does not offer predictive insight into how the model will behave in unseen scenarios.
The psychology concept of attention involves complex processes, whereas the attention mechanism in models is a deterministic computation of similarity scores.
Without a rigorous mapping between theoretical constructs and model components, these psychology references risk becoming superficial narratives—appealing and intuitive, but ultimately can not guide future model development or evaluation.
Interdisciplinary research should place greater emphasis on theoretical modeling in the early stages, encouraging the integration of clear psychological hypotheses during the experimental design phase, rather than retrofitting existing theories only at the analysis stage.

At the methodological level, the operationalization of psychological concepts in AI research also faces challenges.
Many studies attempt to simulate psychological phenomena by constructing proxy tasks; however, these tasks are often unvalidated and lack construct validity.
We argue that AI research should draw on psychology's strong emphasis on measurement validity and experimental control by incorporating more systematic psychological methods into task design and data interpretation. 
For example, collaborating with psychologists to design experiments, using standardized scales, and reporting the psychological validity of tasks are all feasible strategies for improving methodological quality.

It is worth noting that the field of HCI has long been a prime example of interdisciplinary research.
This field not only values theory-driven research design but also emphasizes methodological diversity and rigor.
Drawing from multiple disciplines, HCI has successfully embedded different knowledge into multiple stages of the research process, such as problem definition, system design, user modeling, and evaluation. 
This systematic approach offers valuable insights for AI research.

\subsection{Toward more responsible interdisciplinary practice}\label{sec:6.3}

To build a stronger and more sustainable bridge between AI and psychology research, we advocate for more responsible interdisciplinary practices.
This is not just about upholding ethical standards; it is also about ensuring scientific rigor.
With that in mind, we offer a few recommendations aimed at promoting clearer standards, more consistent methods, and deeper collaboration across disciplines.

\paragraph{Theoretical accountability.}

When drawing on psychology theories, researchers should clearly explain where these theories come from, what assumptions they rest on, and how far they can reasonably be applied to avoid misinterpretation or conceptual reconstruction.
Furthermore, competing theoretical perspectives should be adequately addressed, with explicit justification for the chosen framework and acknowledgment of its limitations.
Such theoretical accountability not only strengthens research transparency but also provides a thoughtful re-evaluation of the extent to which psychology theories can be meaningfully applied in AI research.

\paragraph{Construct operationalization.}

Interdisciplinary research works best when there is a clear and defensible connection between psychological concepts and the technical tasks. 
We suggest starting with standardized, widely accepted measurement tools from psychology whenever possible. 
If you need to design custom tasks, it is important to clearly explain how those tasks reflect the underlying psychological construct and to back that explanation with theory. 
When appropriate, multiple forms of measurement (e.g., behavioral data, linguistic outputs, and subjective ratings) should be employed to ensure a comprehensive evaluation of the construct.

\paragraph{Collaborative parity.}

Interdisciplinary collaboration should not be about one field just doing what another wants. 
Instead, it really needs to be built on a foundation of mutual respect, where different disciplines are truly co-creating something new.
To make this happen, we think it is important to encourage things like joint authorship across disciplines, working together from the very beginning to formulate research questions, and making sure we weave in a variety of analytical perspectives.

\paragraph{Open interdisciplinary infrastructure.}

We advocate for the development of open knowledge infrastructures that support interdisciplinary collaboration. 
Examples include reusable datasets on psychology constructs and measurement methods, case templates for interdisciplinary research, and cross-referencing maps. 
Such resources would lower the barriers to collaboration, enhance the quality of research, and foster the accumulation and transmission of knowledge across disciplinary communities.

In sum, responsible interdisciplinary practice is not a matter of occasional collaboration, but rather a sustainable and institutionalized research framework. 
By strengthening theoretical accountability, standardizing how we use constructs, and promoting equitable collaboration and shared infrastructure, we can achieve a deeper, more trusting integration of AI and psychology research.

\subsection{Limitations and future directions}\label{sec:6.4}

Although we aim to provide a thorough survey of how psychology research is cited and integrated in AI research, several limitations should be acknowledged, as they may affect the broader applicability and interpretive depth of our findings.

First, in terms of the time range, our analysis focuses on published CS research between December 2022 and March 2025, primarily reflecting the short-term impact and initial integration of psychology research into LLM research.
Given that interdisciplinary influences often exhibit a time lag, the deeper transformation of psychology theories, methodological integration, and practical impact may still be in the early stages.
As a result, our findings may underestimate the long-term knowledge diffusion and paradigm-shifting role of psychology in LLM research.

Secondly, in terms of the analyzed data, we primarily examined English-based papers from top AI conferences, which may have led to the omission of relevant work from other important fields. 
Additionally, differences in database indexing mechanisms and citation formats may have caused some psychology sources to be overlooked, potentially resulting in a possible underrepresentation of psychology sources.

Purely exploring the integration of AI research and psychology research from the citation-based perspective might be a limitation.
Future studies could extend this work by moving beyond citation-based analyses to explore the actual processes of interdisciplinary collaboration between AI and psychology. 
This may include examining how knowledge is negotiated across disciplinary boundaries, how research teams are structured, and which collaboration mechanisms are most effective. 
Such qualitative insights would complement the current study’s bibliometric approach and provide a more comprehensive understanding of how these fields interact in practice.

\section{Conclusion}\label{sec:7}

This work contributes to a growing science of science approach to understanding how interdisciplinary knowledge circulates, mutates, and influences AI development.
By identifying the domains and dynamics of psychology research influence in LLM research, we aim to provide not only a descriptive map but also a normative guide: showing how psychology research is most productively integrated, where misuse arises, and how better practices can be cultivated. 
As AI systems become increasingly embedded in the fabric of society, the importance of methodological pluralism, conceptual clarity, and cross-disciplinary rigor will only grow.
Psychology has helped us understand human intelligence; with care and collaboration, it may also help us build AI more wisely.

\begin{ack}
This research project is partially sponsored by the Microsoft Accelerate Foundation Models Research (AFMR) grant program.
\end{ack}

\clearpage

\small
\bibliographystyle{elsarticle-harv}
\bibliography{references}

\clearpage
\newpage

\appendix

\section{Instructions for GPT}\label{app:instruction}
\begin{figure}
    \centering
    \includegraphics[width=\linewidth]{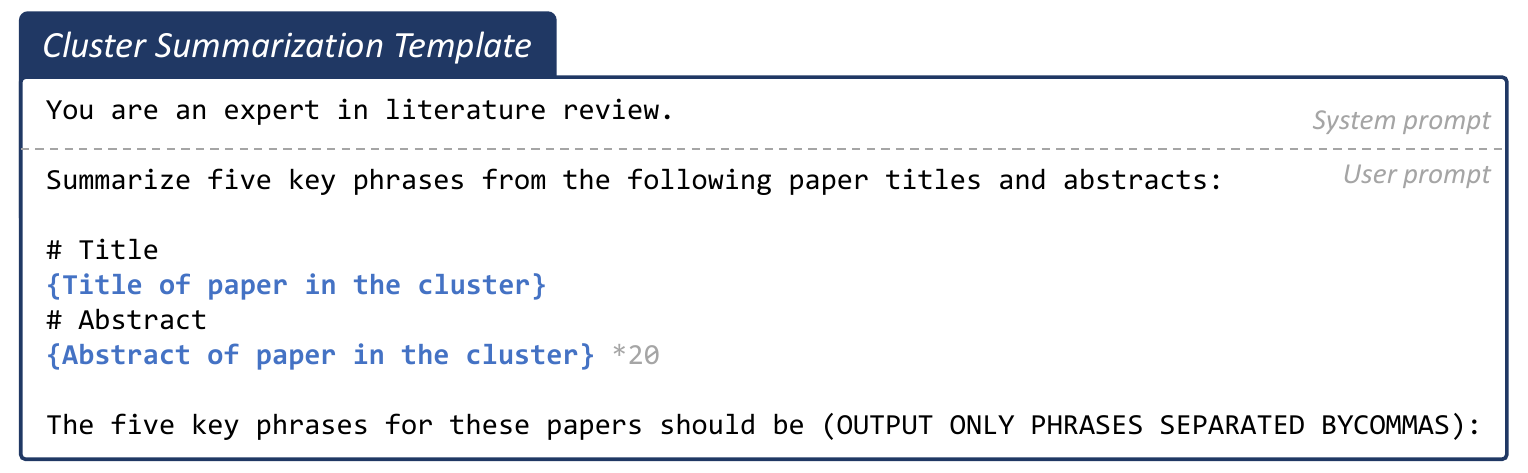}
    \caption{Instruction Template for GPT-4o to Summarize Key Phrases in a Paper Cluster.}
    \label{fig:inst_summarize}
\end{figure}

\begin{figure}
    \centering
    \includegraphics[width=\linewidth]{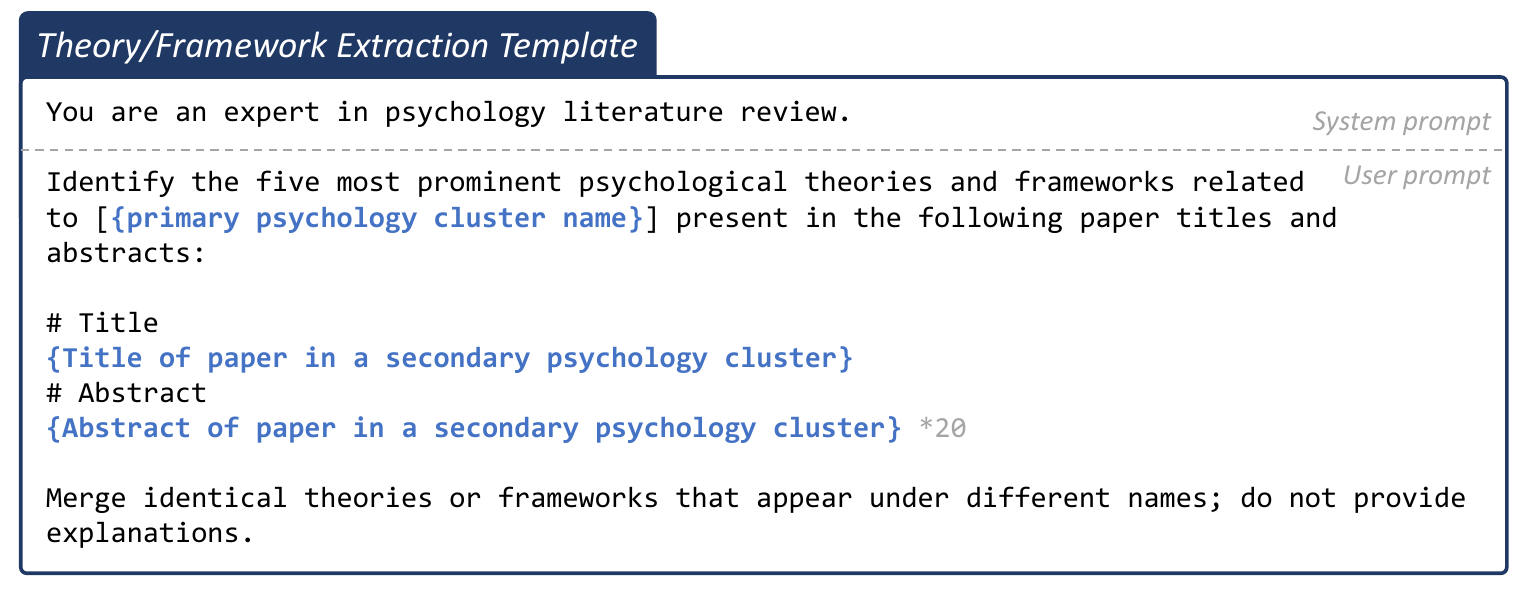}
    \caption{Instruction Template for GPT-4.1 to Extract Candidate Psychology Theories and Frameworks from a Secondary Psychology Cluster.}
    \label{fig:inst_extract}
\end{figure}

\begin{figure}
    \centering
    \includegraphics[width=\linewidth]{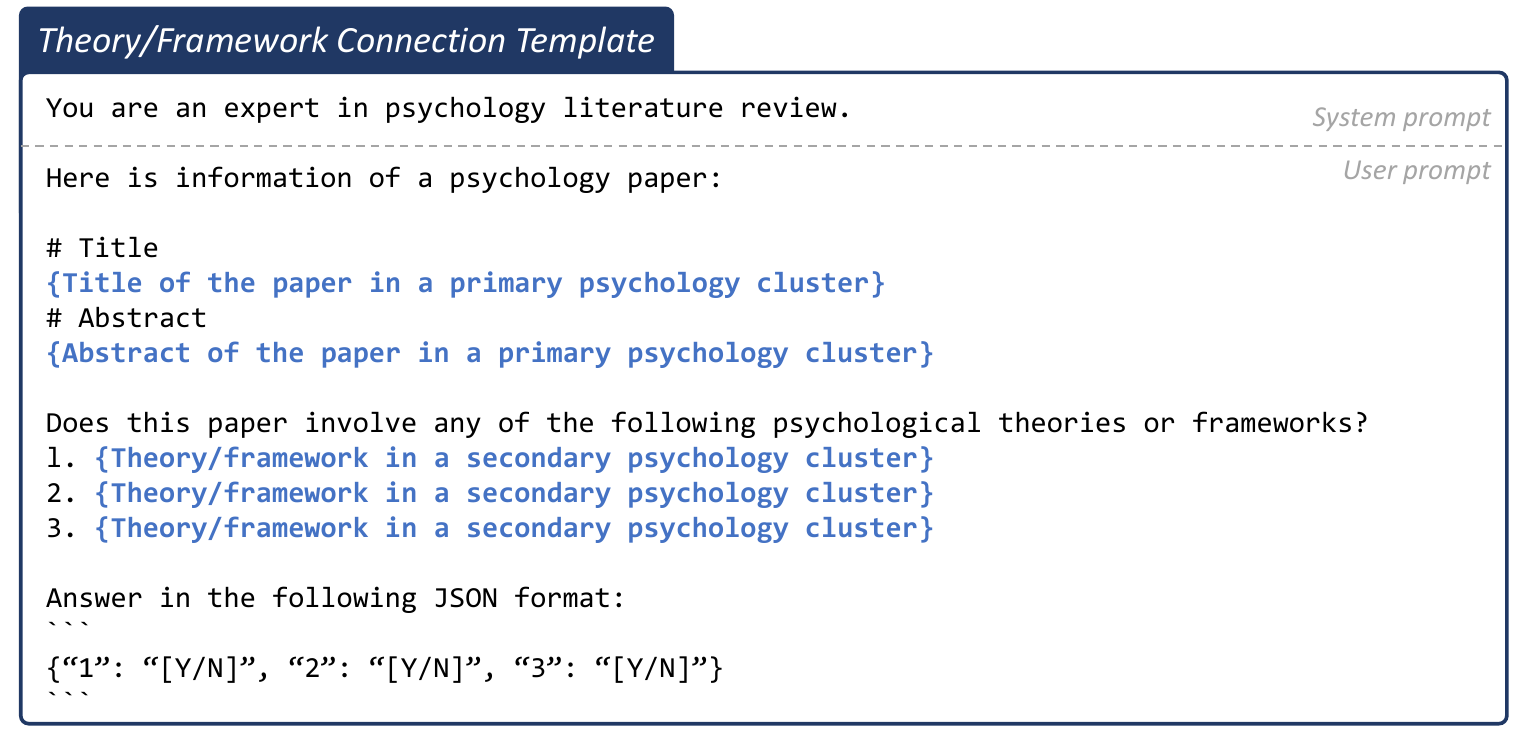}
    \caption{Instruction Template for GPT-4.1 to Link a Psychology Paper with Theories and Frameworks in a Secondary Psychology Cluster.}
    \label{fig:inst_connect}
\end{figure}

In the citation analysis, we used GPT-4o to derive cluster topics and GPT-4.1 to extract and connect psychology theories and frameworks, as mentioned in §\ref{sec:3.2} and §\ref{sec:3.3}, respectively. The instruction templates are provided in Fig.~\ref{fig:inst_summarize}, Fig.~\ref{fig:inst_extract}, and Fig.~\ref{fig:inst_connect}.

\end{document}